\documentclass[lettersize,journal]{IEEEtran}
\usepackage{amsmath,amsfonts}
\usepackage{algorithmic}
\usepackage{algorithm}
\usepackage{array}
\usepackage[caption=false,font=normalsize,labelfont=sf,textfont=sf]{subfig}
\usepackage{textcomp}
\usepackage{stfloats}
\usepackage{url}
\usepackage{verbatim}
\usepackage{graphicx}
\usepackage{cite}
\usepackage{float}
\usepackage{booktabs}
\usepackage{multicol}
\usepackage{multirow}
\usepackage{diagbox}
\usepackage{colortbl}
\usepackage{xcolor}
\usepackage{amssymb}
\newcommand{\figref}[1]{Figure~\ref{#1}}
\newcommand{\tabref}[1]{Table~\ref{#1}}

\newcommand{\equref}[1]{Eq.~(\ref{#1})}

\definecolor{revisionred}{RGB}{0,0,0}
\newif\ifshowchanges
\showchangestrue

\DeclareRobustCommand{\rev}[1]{%
  {\ifshowchanges\color{revisionred}\fi#1}%
}

\newenvironment{revblock}
  {\ifshowchanges\color{revisionred}\fi}
  {}

\begin{document}

\title{TimeMachine: Fine-Grained Facial Age Editing with Identity Preservation}

\author{Yilin Mi\textsuperscript{*}, 
    Qixin Yan\textsuperscript{*}, 
    Zheng-Peng Duan,
    Chunle Guo, 
    Hubery Yin,
    Hao Liu,
    Chen Li,
    Jing Lyu,
    and Chongyi Li

\thanks{This work was supported in part by the Major Science and Technology Project on Artificial Intelligence of Tianjin (No. 25ZXRGGX00290), the Tianjin Natural Science Foundation Project (25ZXRGGX00290, 24JCJQJC00020, 25JCQNJC01390), and the Fundamental Research Funds for the Central Universities (Nankai University, 63243143, 63253223, 63253219).  (\emph{Corresponding author: Chongyi Li.})}
    \thanks{Yilin Mi and Qixin Yan contributed equally to this work. }
    \thanks{
    Yilin Mi and Zheng-Peng Duan are with VCIP,
    College of Computer Science, Nankai University, Tianjin 300350,
    China (e-mail: yilinmi@mail.nankai.edu.cn; adamduan0211@gmail.com) .}
    \thanks{
    Qixin Yan, Hubery Yin, Hao Liu, Chen Li, and Jing Lyu are with Weixin, Tencent, China (e-mail: qixinyan@tencent.com; hubery@tencent.com; leweshaoliu@tencent.com; chaselli@tencent.com; eckolv@tencent.com)
    }
    \thanks{
    Chunle Guo and Chongyi Li are with VCIP, College of Computer Science,
    Nankai University, Tianjin 300350, China, and also with the Nankai International Advanced Research Institute (NKIARI), Futian, Shenzhen 518045,
    China (e-mail: guochunle@nankai.edu.cn; lichongyi@nankai.edu.cn). 
}
}

\maketitle

\begin{abstract}
With the advancement of generative models, facial image editing has made significant progress. 
However, achieving fine-grained age editing while preserving personal identity remains a challenging task.
\rev{In this paper, we propose \textbf{TimeMachine}, a unified diffusion-based framework for accurate facial age editing with identity preservation.}
\rev{TimeMachine employs fine-grained age conditions and a Multi-Cross-Attention module to decouple the dedicated age and identity conditions.}
\rev{This branch-wise design enables controllable age transformation while maintaining identity consistency.}
\rev{Furthermore, we introduce an Age Classifier Guidance (ACG) module that predicts age embeddings from latent denoising features without reconstructing images during training.}
\rev{This lightweight training-only predictor provides additional age-related supervision and is removed during inference.}
\rev{To support large-scale training, we construct \textbf{HFFA} (High-quality Fine-grained Facial-Age), a million-scale collection of high-resolution face images annotated with captions, identity embeddings, MiVOLO-estimated numerical ages, and age embeddings.}
\rev{Extensive experiments across five public face datasets demonstrate that TimeMachine achieves superior age accuracy and identity preservation relative to the evaluated baselines, with consistent conclusions across five age estimators. The project page is available at https://miyilin2003.github.io/TimeMachine}
\end{abstract}


\section{Introduction}
\IEEEPARstart{H}{uman} face serves as a powerful canvas, conveying not only identity but also other attributes such as age, emotion, and health.
Recent advancements in image generation have led to significant progress in facial image analysis and editing.
Among these techniques, 
age manipulation, covering both age progression and regression,
has attracted increasing attention.
It allows us to digitally alter the perceived age of individuals in images and videos,
offering many practical and entertaining uses.

However, fine-grained facial age editing remains challenging because age-related appearance changes must be introduced without disrupting instance-specific identity characteristics.
Earlier GAN-based approaches commonly rely on age-group conditioning or age-related latent directions~\cite{goodfellow2014generative}, which may limit precise numerical age control and identity preservation.
\begin{revblock}
Diffusion-based personalized generation methods provide stronger generative priors and effective identity conditioning, but they are primarily designed to reproduce a reference identity rather than explicitly coordinate it with a fine-grained target-age condition~\cite{li2024photomaker,wang2024instantid,ye2023ip}.
Consequently, textual age prompts or generic image conditioning alone may provide insufficient numerical age control or introduce unintended changes to non-target facial characteristics.
Building on established projection-and-cross-attention conditioning, TimeMachine constructs dedicated identity, averaged age-codebook, and numerical age-text conditions and processes them through semantically assigned attention branches.
This branch-wise architectural and functional separation supports controllable age transformation while preserving the identity characteristics of the input subject.
These conditions have complementary roles: the identity embedding carries subject-specific information, the averaged codebook provides a shared age representation, the numerical age phrase specifies the target value, and the original text condition supplies appearance and scene context.
\end{revblock}

\begin{figure}[t]
    \centering
    \includegraphics[width=\linewidth]{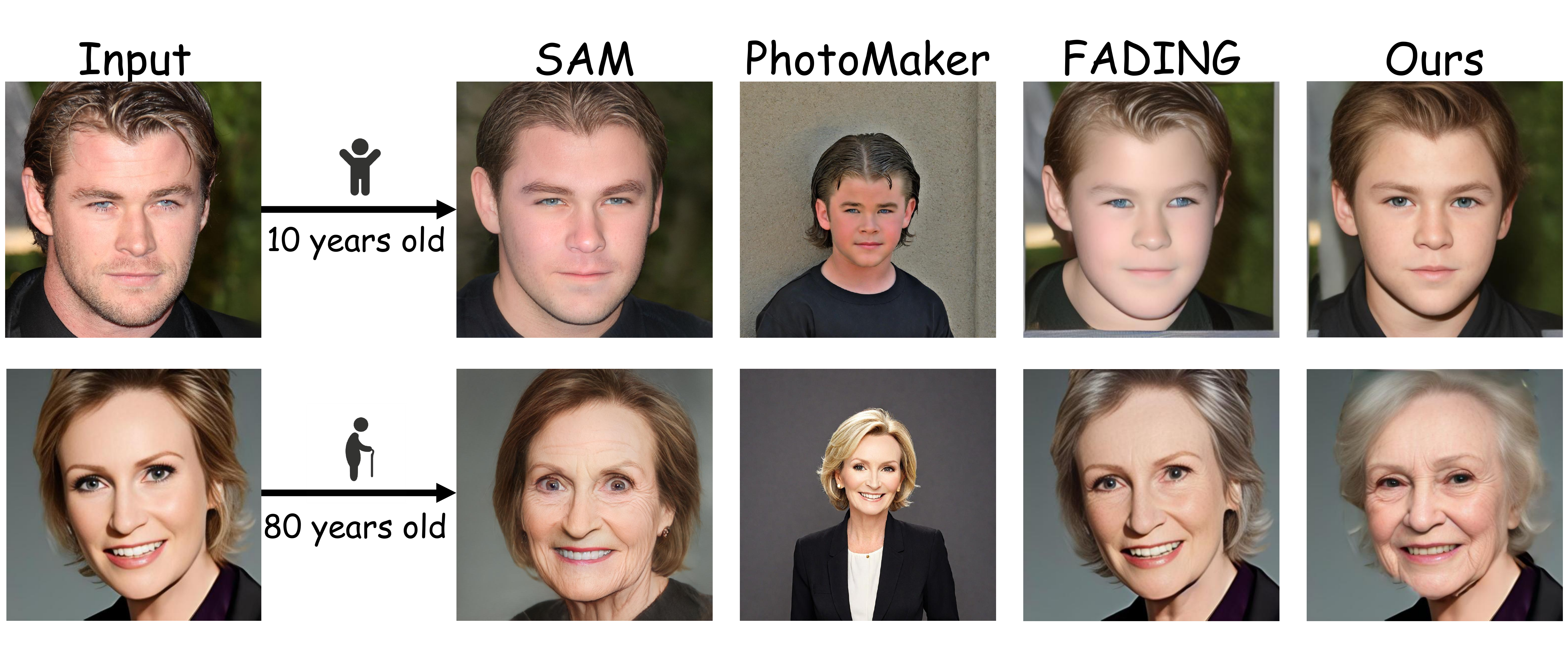}
\caption{We compare TimeMachine with several state-of-the-art face editing approaches, including GAN-based SAM~\cite{alaluf2021only} and diffusion-based PhotoMaker~\cite{li2024photomaker} and FADING~\cite{chen2023face}. Given the same input identity and target age, the examples illustrate differences in age transformation, identity preservation, and visual quality. Aggregate quantitative comparisons are reported in Section~IV. (Zoom in for the best view.)}
    \label{fig:teaser}
    \vspace{-15pt} 
\end{figure}

In this paper, we address the challenge of \rev{separating fine-grained age control from identity preservation} and propose a unified framework for facial age editing. Our contributions include the following four aspects.
\textbf{First}, \rev{building on projection-and-cross-attention conditioning,} we formulate Multi-Cross-Attention for fine-grained facial age editing \rev{using semantically assigned condition branches with branch-specific projections and attention normalization.}
\textbf{Second}, we introduce a \rev{lightweight training-only ACG module} that aligns latent denoising features with \rev{target age embeddings} without requiring image reconstruction or pixel-space age regression.
\textbf{Third}, we construct a numerical age codebook by averaging the age embeddings assigned to each age, \rev{providing a shared age representation designed to reduce sample-specific identity variation in the age condition.}
\textbf{Fourth}, we present HFFA, a million-scale high-resolution facial-age collection \rev{with rich captions, identity embeddings, and MiVOLO-estimated numerical age labels and embeddings.}
\rev{\textit{Extensive evaluations across five public face datasets demonstrate improved age accuracy and strong identity preservation relative to the evaluated baselines.}}

\section{Related Work}

\begin{figure*}[t]
    \centering
    \includegraphics[width=0.9\textwidth]{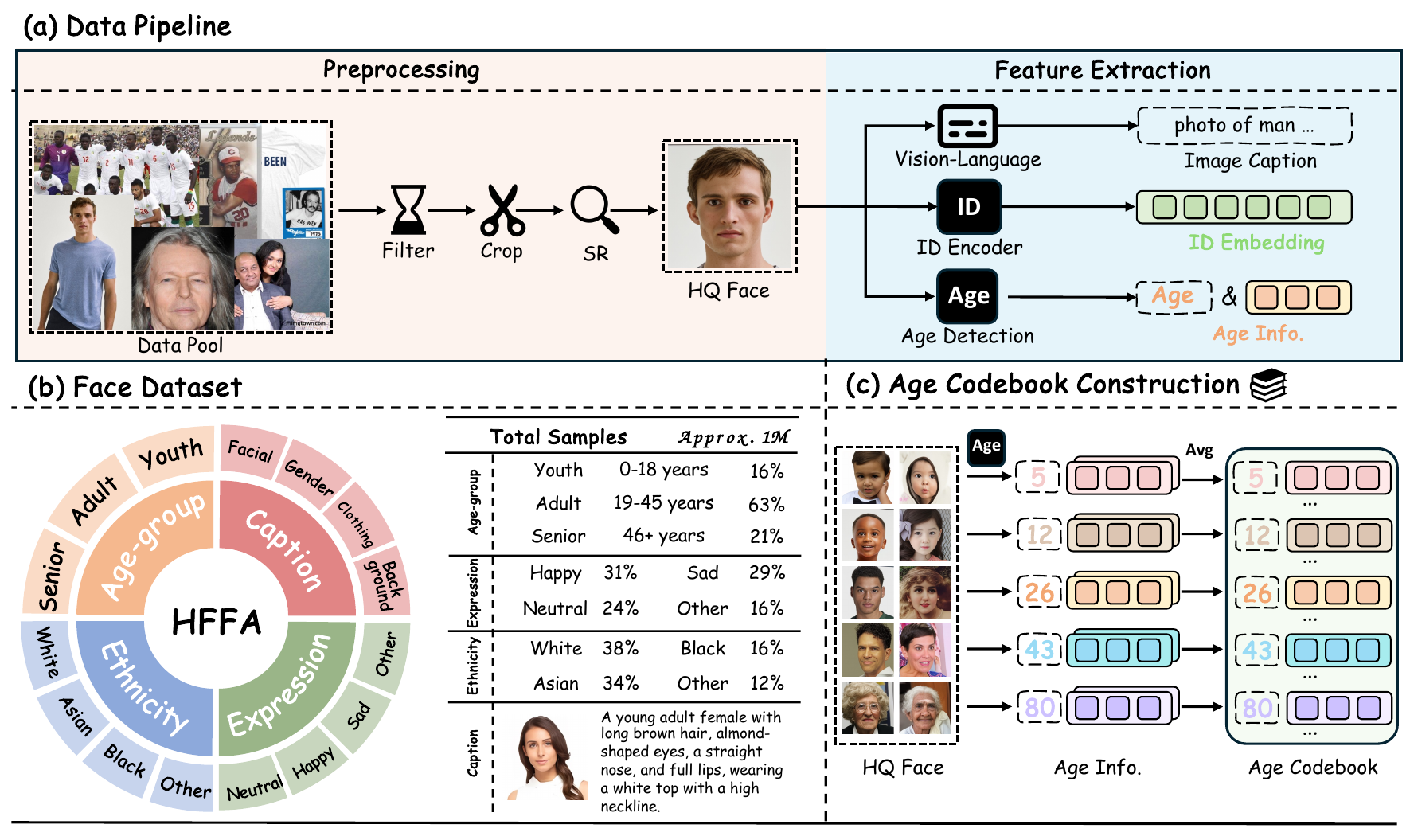}
        \vspace{-10pt} 
\caption{\textbf{HFFA} construction pipeline. Each facial image undergoes filtering, cropping, resizing, and super-resolution before a pretrained age estimator, a vision-language model, and a face representation extractor produce the corresponding annotations. Embeddings assigned to the same numerical age are averaged to construct an \textbf{age codebook} that provides shared age representations \rev{while reducing sample-specific variation}.}
    \label{fig:datapool}
        \vspace{-10pt} 
\end{figure*}

\begin{figure*}[htb]
    \centering
    \includegraphics[width=\textwidth]{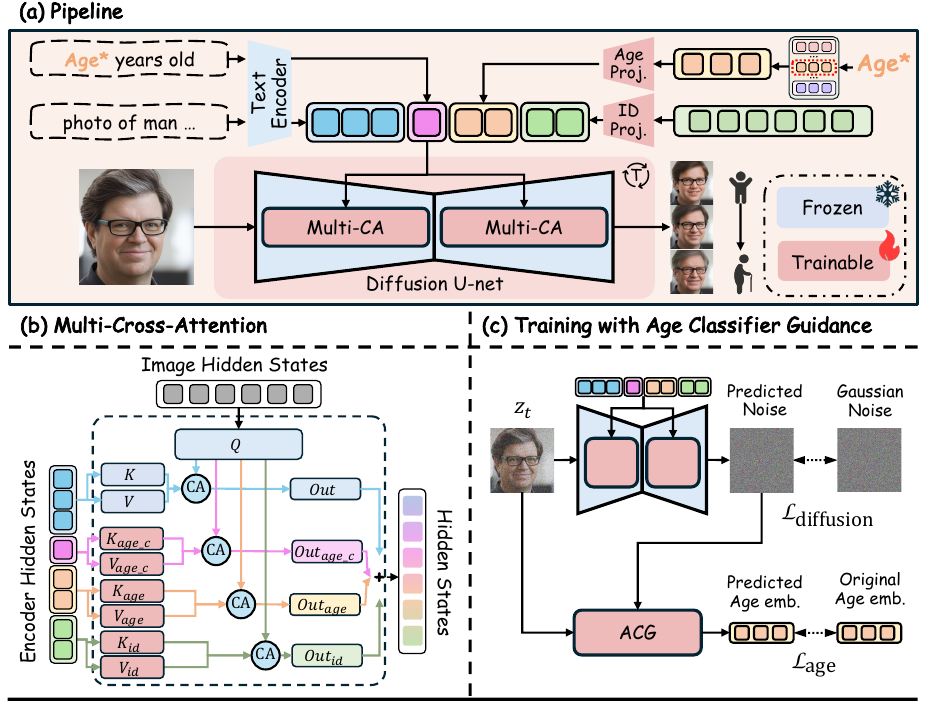}
        \vspace{-15pt} 
\caption{Overview of the TimeMachine pipeline. (a) Age and identity conditions guide the diffusion model toward fine-grained age transformation with identity preservation. \rev{(b) Multi-Cross-Attention processes text, numerical age-text, identity, and age conditions through separate attention branches. (c) The training-only ACG module predicts age embeddings from latent denoising features and provides additional age-related supervision without pixel-space reconstruction.}}
    \label{fig:pipeline}
        \vspace{-15pt} 
\end{figure*}

\subsection{GAN-Based Facial Age Editing Methods}

Early facial age-editing methods predominantly rely on GANs~\cite{goodfellow2014generative}, using either age-conditioned generation~\cite{antipov2017face,wang2018face,zhang2017age} or age-related directions in pretrained latent spaces~\cite{alaluf2021only,or2020lifespan,harkonen2020ganspace,shen2020interpreting}.
Subsequent studies improve continuous age control, reconstruction fidelity, and identity preservation~\cite{li2021continuous,gomez2022custom,yao2021high}.
Nevertheless, age-related latent manipulation may also alter correlated facial characteristics~\cite{harkonen2020ganspace,shen2020interpreting,tov2021designing}, and many methods provide age-group or latent-direction control rather than direct fine-grained numerical-age conditioning.

\subsection{Diffusion Models in Face Editing}
Diffusion models~\cite{ho2020denoising,ramesh2022hierarchical,avrahami2023blended,meng2021sdedit,nichol2021glide,hertz2022prompt,mokady2023null,rombach2022high,saharia2022photorealistic,dhariwal2021diffusion} have achieved strong performance in image generation tasks, including facial synthesis and editing. These models generate images by iteratively denoising random noise, enabling them to capture complex data distributions and fine-grained facial details. Their strong generative priors make them particularly suitable for modeling the gradual texture and structural changes involved in facial age transformation~\cite{chen2023face}. However, general-purpose diffusion editing methods~\cite{couairon2022diffedit,kawar2023imagic} are not specifically designed to coordinate identity preservation with fine-grained numerical-age control.

Recent identity-preserving personalized generation methods~\cite{ye2023ip,li2024photomaker,wang2024instantid,guo2024pulid} incorporate reference-image or face-identity conditions into the denoising process and substantially improve subject fidelity. Fine-grained facial age editing additionally requires explicit coordination between the identity condition and the target-age condition, which is not the principal focus of these methods. IP-Adapter~\cite{ye2023ip} introduces decoupled cross-attention for image-prompt conditioning. TimeMachine builds upon this established projection-and-cross-attention principle and jointly constructs and optimizes four task-specific condition branches for fine-grained facial age editing.

\section{Methodology}

\begin{figure*}[htb]
    \centering
    \includegraphics[width=\textwidth]{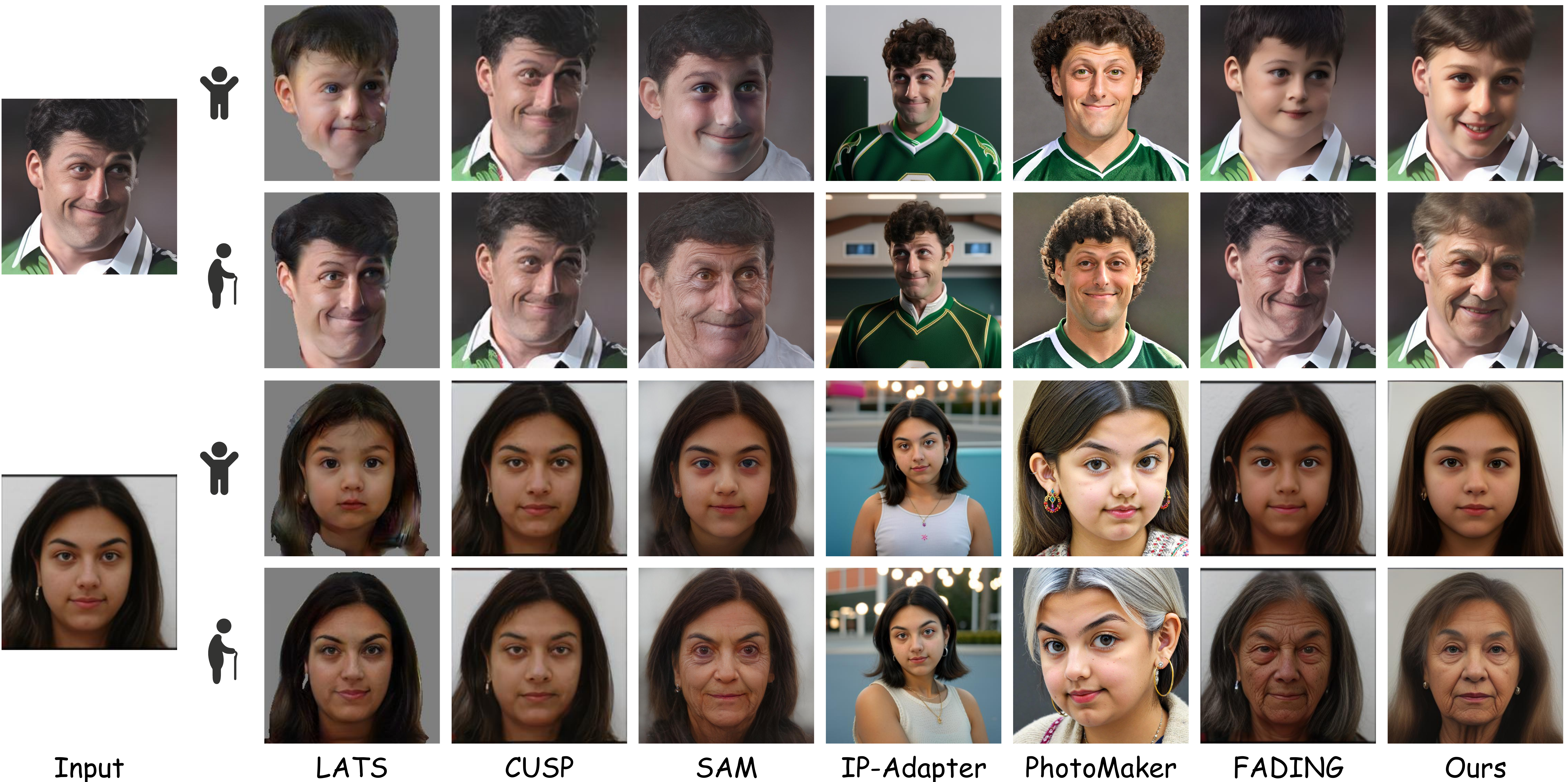}
\caption{Qualitative comparison with state-of-the-art methods. We conduct extreme age editing generation tests on each identity, showing results at age 10 (top row) and age 80 (bottom row). Our model effectively performs age editing while preserving identity information.}
    \label{fig:demo_10_and_80}
        \vspace{-15pt} 
\end{figure*}

\subsection{Preliminaries}
Stable Diffusion (SD)~\cite{rombach2022high} builds on Denoising Diffusion Probabilistic Models~\cite{ho2020denoising} and performs the denoising process in a learned latent space with CLIP-based text conditioning~\cite{radford2021learning}.
Given a clean latent $z_0$, text embedding $c$, sampled timestep $t$, and Gaussian noise $\epsilon$, the denoising network is optimized using the standard noise-prediction objective:
\begin{equation}
\mathcal{L}_{{diffusion}} = \mathbb{E}_{z_0, c, \epsilon, t} \left[ \| \epsilon - \epsilon_\theta( z_t, t, c) \|_2^2 \right],
\label{eq:sd_loss}
\end{equation}
where $z_t$ is the noised latent code at timestep $t$, $\epsilon$ is the noise added to the latent $z_t$, and $\epsilon_\theta$ is the text-conditioned denoising network.

\subsection{Training Data Pool Construction}
\label{sec:DataPool}

High-quality training data is important for fine-grained facial age editing. Public datasets with \rev{chronological or apparent age annotations}, such as MORPH~\cite{ricanek2006morph} and AgeDB~\cite{AGEDB8014984}, are relatively limited in scale, whereas web-scale face collections~\cite{zheng2022general} generally \rev{lack fine-grained age annotations} and exhibit substantial variation in image quality.

To address this issue, as visualized in \figref{fig:datapool}, we construct \textbf{HFFA} (High-quality Fine-grained Facial-Age), a million-scale collection in which each image is accompanied by a text caption ($c$), an identity embedding ($e_{id}$), an age embedding ($e_{age}$), and a \rev{MiVOLO-estimated numerical age} ($age$). HFFA combines \rev{fine-grained model-estimated age annotations} with high visual quality, supporting the learning of facial details required for age transformation.

In the final processed HFFA collection, \rev{96\% of the facial images are derived from LAION-Face \cite{zheng2022general}, while the remaining 4\% constitute a self-collected supplementary facial-image subset obtained from publicly accessible web sources. This supplementary subset is introduced primarily to alleviate the underrepresentation of Asian facial images in LAION-Face.} Both portions undergo the same preprocessing pipeline, including single-face filtering, facial-region cropping and alignment, and image enhancement.

We first detect and extract single-face regions from the raw images.
Each detected face is aligned, resized, and tightly cropped around the facial region.
We then apply CodeFormer~\cite{zhou2022towards} to improve the resolution and facial details of the cropped images.
The resulting collection contains visually consistent single-face images suitable for training the facial age-editing model.

For each processed image, textual descriptions are generated using the Qwen2.5-VL-72B-Instruct~\cite{bai2025qwen25vl}, which provides natural language captions capturing facial attributes, context, and appearance. To obtain identity information, we use Antelopev2~\cite{InsightFace_ModelZoo} to extract identity embeddings $e_{id}$. For each processed image $x_j$, the \rev{frozen} MiVOLO model~\cite{kuprashevich2023mivolo} produces a \rev{continuous} numerical age estimate $\tilde{a}_j$ from its output layer and a \rev{384-dimensional} age embedding $e_{\mathrm{age}}^j$ from its penultimate age-representation layer. \rev{The continuous estimate is rounded to the nearest integer, yielding numerical age labels spanning 1 to 85 in the processed HFFA corpus.} These \rev{model-estimated labels} provide a consistent fine-grained age index for age-codebook construction and model training. 

Because an individual age embedding may retain sample-specific information, we average the embeddings assigned to each numerical age to construct an \textbf{age codebook}. The resulting shared age representations are \rev{designed to reduce sample-specific identity variation in the age condition} and facilitate branch-wise age-identity decoupling during training.

\subsection{Age Editing with Multi-Cross-Attention}

For each editing sample, TimeMachine constructs four condition groups: the Qwen-generated caption $c$, an explicit numerical age phrase $c_{\mathrm{age}}$, the Antelopev2 identity embedding $e_{\mathrm{id}}$ of the input face, and the averaged age-codebook entry $e_{\mathrm{code}}^{a}$ corresponding to target age $a$.
The identity and age representations are mapped through dedicated projection modules, while the two textual conditions are encoded separately and routed to their corresponding Multi-CA branches.

\subsubsection{Condition Projection Module}

\leavevmode
\rev{Following the projection-and-cross-attention principle of IP-Adapter \cite{ye2023ip}}, we use separate learnable projection modules for the identity and age conditions. Antelopev2 extracts the instance-level identity embedding $e_{\mathrm{id}}$, which is mapped by the image projection module to identity tokens $\hat e_{\mathrm{id}}$. For a target age $a$, the corresponding \rev{averaged codebook entry $e_{\mathrm{code}}^{a}$} is mapped by the age projection module to age tokens $\hat e_{\mathrm{age}}$. The Qwen-generated caption $c$ is processed as the original Stable Diffusion text condition, whereas an explicit numerical age phrase $c_{\mathrm{age}}$, such as ``25 years old,'' is \rev{encoded separately and routed to the age-text branch}.

The four branches have complementary roles. The ID branch supplies \rev{instance-level identity information}; the age and age-text branches provide \rev{the averaged age representation and exact numerical target, respectively}; and the text branch supplies \rev{facial appearance and scene semantics}. \rev{TimeMachine implements branch-wise decoupling by constructing the four condition groups separately and processing them with branch-specific key/value projections and attention normalizations; the identity and age-codebook embeddings additionally pass through dedicated projection modules.}

For implementation, \rev{the four token groups are concatenated before entering the attention processor and partitioned into their corresponding groups inside each Multi-CA processor.} Each group is then processed by its branch-specific key/value projections and \rev{softmax normalization in \equref{eq:multica}}, while \rev{all trainable projection and attention parameters are jointly optimized for the facial age-editing objective.}

\subsubsection{Decoupled Multi-Cross-Attention}
\leavevmode

Multi-CA is integrated into each UNet block that contains cross-attention and processes four condition types through parallel branches.
We denote the condition tokens by $C_i$, where $i\in\{\mathrm{text},\mathrm{ID},\mathrm{age},\mathrm{age\text{-}text}\}$.

\begin{figure*}[htb]
    \centering
    \includegraphics[width=\textwidth]{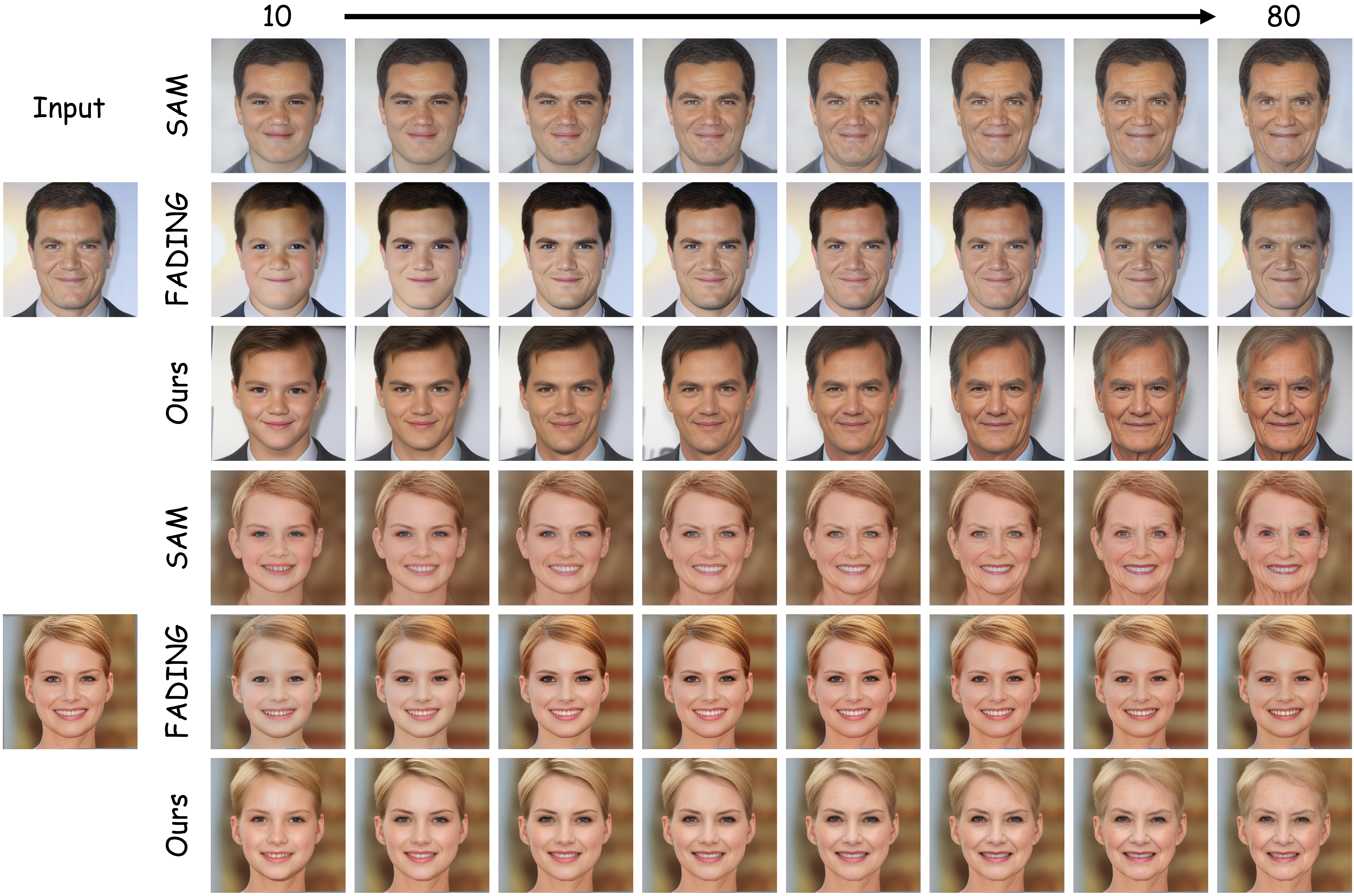}
    \caption{Representative progressive-age comparison among SAM, FADING, and TimeMachine over target ages from 10 to 80. The displayed sequences illustrate differences in age-transition strength, visual consistency, and the preservation of subject-specific facial characteristics.}
    \label{fig:demo_10_to_80}
        \vspace{-3pt} 
\end{figure*}

\begin{revblock}
Given the current latent feature $H$, all branches derive their query from the same latent representation, while each condition branch uses its own key and value projections and an independent softmax normalization:
\begin{equation}
O_i=\operatorname{Attn}(Q,K_i,V_i),\quad
H'=\left(\sum_i\lambda_iO_i\right)W_O.
\label{eq:multica}
\end{equation}
Here, $Q=HW_Q$, $K_i=C_iW_K^i$, $V_i=C_iW_V^i$, and $\operatorname{Attn}(Q,K,V)=\operatorname{Softmax}(QK^\top/\sqrt d)V$.
The matrices $W_K^i$ and $W_V^i$ are branch-specific projections, whereas $W_O$ is shared.
Because every $O_i$ is computed from the same query $Q$, the branch outputs have aligned spatial and channel dimensions before fusion.
The weighted sum in \equref{eq:multica} fuses these query-aligned latent attention features within each cross-attention block.
Independent normalization also prevents heterogeneous condition tokens from competing within a single softmax.
\end{revblock}

\begin{figure*}[h]
    \centering
    \includegraphics[width=\textwidth]{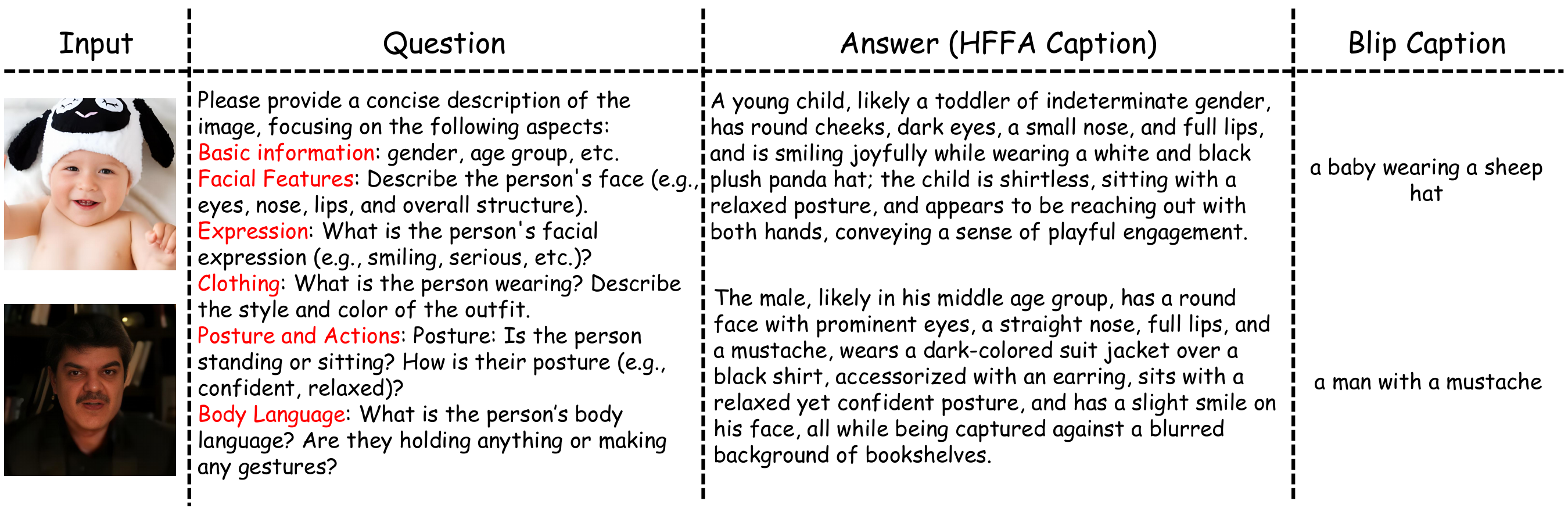}
    \caption{Overview of the captioning process for our dataset. We use Qwen2.5-VL-72B-Instruct to generate captions by asking specific questions about the images, and then use the obtained responses as the captions. The results are compared with those generated by the BLIP model.}
    \label{fig:caption}
\end{figure*}

\subsection{Training with Age Classifier Guidance}

Standard diffusion training supervises noise prediction, while ACG adds an explicit age-embedding objective to intermediate denoising features during Stage II.
ACG operates only during training and is removed at inference.
\begin{revblock}
MiVOLO is used during HFFA annotation to construct the clean-image age-codebook targets.
For a training sample assigned to integer age \(a\), ACG uses the corresponding averaged codebook entry
\(c_a\equiv e_{\mathrm{code}}^{a}\in\mathbb{R}^{384}\) as its detached target.
Given the current noisy latent \(z_t\), timestep \(t\), and the noise residual \(\hat{\epsilon}_{\theta}\) predicted by the denoising network, ACG directly predicts a 384-dimensional age embedding:
\begin{equation}
\hat{\epsilon}_{\theta}=\epsilon_{\theta}(z_t,t,\mathcal{C}),\quad
\hat e_{\mathrm{age}}=G_{\phi}(z_t,\hat{\epsilon}_{\theta},t)\in\mathbb{R}^{384}.
\label{eq:acg_predict}
\end{equation}
where \(\mathcal{C}\) denotes the complete set of conditioning signals.
ACG therefore maps \((z_t,\hat{\epsilon}_{\theta},t)\) directly to the target age-embedding space.
\end{revblock}
Its computation remains entirely in latent feature space and requires neither clean-latent estimation nor image decoding.

\begin{revblock}
\textbf{Timestep-aware dual-stream predictor.}
We encode timestep \(t\) with a 128-dimensional sinusoidal embedding and a two-layer MLP to obtain a 256-dimensional feature \(e_t\).

\begin{samepage}
Two independent convolutional encoders process \(z_t\) and \(\hat{\epsilon}_{\theta}\):
\begin{equation}
e_t=E_t(t),\quad
(f_z,f_{\epsilon})=\bigl(E_z(z_t,e_t),E_{\epsilon}(\hat{\epsilon}_{\theta},e_t)\bigr).
\label{eq:acg_encoders}
\end{equation}
\end{samepage}

Each encoder contains a convolutional stem and three timestep-conditioned residual downsampling blocks, with channel dimensions of 32, 64, 128, and 256.
Global average pooling and layer normalization produce the 256-dimensional features \(f_z\) and \(f_{\epsilon}\).

The noisy-latent feature \(f_z\) and timestep feature \(e_t\) produce a bounded channel-wise gate over \(f_{\epsilon}\):
\begin{equation}
g_t=1+0.5\tanh\!\left(W_g\operatorname{Cat}(f_z,e_t)\right),\quad
h_{\mathrm{ACG}}=\operatorname{LN}\!\left(f_{\epsilon}\odot g_t\right).
\label{eq:acg_gate}
\end{equation}
Consequently, each element of \(g_t\) lies in \([0.5,1.5]\), allowing the noisy-latent and timestep features to modulate the predicted-noise representation within a bounded range.
A two-layer prediction head with dimensions \(256\!\rightarrow\!512\!\rightarrow\!384\) maps \(h_{\mathrm{ACG}}\) to the normalized prediction \(\hat e_{\mathrm{age}}\).

ACG is optimized by cosine similarity, and its loss is added to the Stage-II denoising objective:
\begin{equation}
\mathcal{L}_{\mathrm{ACG}}=1-\cos(\hat e_{\mathrm{age}},c_a),\quad
\mathcal{L}_{\mathrm{total}}=\mathcal{L}_{\mathrm{diffusion}}+0.05\,\mathcal{L}_{\mathrm{ACG}}.
\label{eq:acg_objective}
\end{equation}
During Stage II, the target \(c_a\) is detached, whereas the gradient through \(\hat{\epsilon}_{\theta}\) is retained.
ACG is removed after Stage II, leaving the inference architecture and sampling procedure unchanged.
\end{revblock}

\section{{Experiments}}

\subsection{More Details of HFFA}

\subsubsection{\rev{Responsible Data Use and Access}}

\begin{revblock}
The supplementary subset was collected from publicly accessible web sources and screened according to the source-use terms available at the time of collection for non-commercial academic research.
The collection and current research use were reviewed under the applicable institutional procedures for publicly accessible data.
The retained records are limited to facial images and model-derived annotations required for training; real names, contact information, user-profile identifiers, source-page URLs, and explicit real-world identity labels are not stored.
Upon formal acceptance, the code and pretrained checkpoint will be publicly released.
We will explore controlled academic access for the subset approved for sharing.
\end{revblock}

\subsubsection{Preprocessing details}

\begin{figure}[h]
    \centering
    \includegraphics[width=\linewidth]{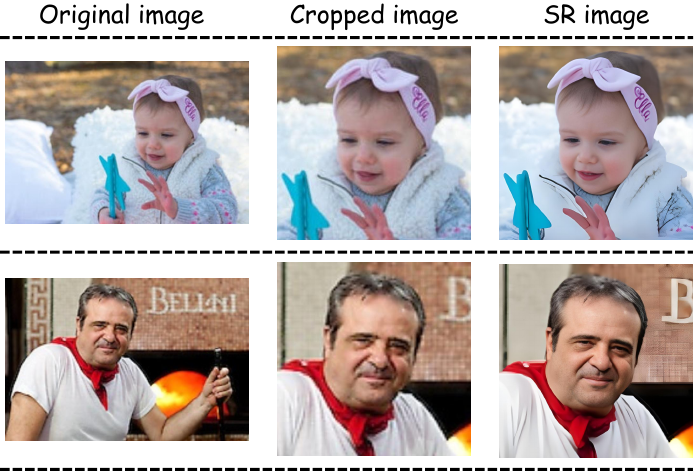}
    \caption{Results of the preprocessing steps, including cropping and super-resolution (SR) enhancement.}
    \label{fig:preprocessing}
\end{figure}

We retain images containing a single detected face that occupies more than 5\% of the image area.
The facial region is then cropped and enhanced using CodeFormer, as illustrated in \figref{fig:preprocessing}.

Qwen2.5-VL-72B-Instruct~\cite{bai2025qwen25vl} is used to generate captions describing visible facial, appearance, and scene information.
The representative comparison in \figref{fig:caption} shows that the Qwen captions capture more of these visible details than BLIP~\cite{li2022blip} in the displayed examples.
\begin{revblock}
These captions provide complementary text context and may contain coarse age-group descriptions, whereas the exact target age is supplied separately through the numerical age-text and age-codebook conditions.
\end{revblock}

\subsubsection{\rev{Cross-Estimator Consistency of HFFA Age Labels}}

\begin{revblock}
To directly assess the consistency of the model-estimated age annotations, we randomly sample 10,000 images from HFFA with a fixed random seed and independently re-estimate their ages using MIVIA SE-ResNet50~\cite{Greco_2022_MIVIA}, Faceptor~\cite{Qin_2024_Faceptor}, Phi-4 Multimodal~\cite{Microsoft_2025_Phi4}, and Qwen3-VL-8B~\cite{Bai_2025_Qwen3VL}.
None of these four evaluators participates in HFFA annotation, age-codebook construction, or TimeMachine training.
All evaluators receive the same bounding-box face crop for each image and follow their official inference protocols.
\end{revblock}

\begin{table}[h]
\centering
\footnotesize
\setlength{\tabcolsep}{2.8pt}
\caption{\rev{Cross-estimator consistency between the MiVOLO-derived HFFA age labels and four independent age evaluators on 10,000 randomly sampled HFFA images. Consensus denotes the per-image median of the four independent predictions.}}
\label{tab:hffa_age_label_consistency}
\begin{tabular}{lccccc}
\toprule
Evaluator &
Valid &
Mean $|\Delta a|$ $\downarrow$ &
$\pm5$ $\uparrow$ &
$\pm10$ $\uparrow$ &
Pearson $r$ $\uparrow$ \\
\midrule
MIVIA     & 10,000 & 5.14 & 59.58\% & 88.27\% & 0.855 \\
Faceptor  & 10,000 & 4.70 & 68.19\% & 88.43\% & 0.830 \\
Phi-4     & 9,998  & 4.58 & 69.74\% & 91.63\% & 0.905 \\
Qwen3-VL  & 10,000 & 5.22 & 63.81\% & 88.53\% & 0.932 \\
\midrule
Consensus & 9,998 & \textbf{3.47} & \textbf{78.28\%} &
\textbf{96.73\%} & \textbf{0.945} \\
\bottomrule
\end{tabular}
\end{table}

\begin{revblock}
The MiVOLO-derived HFFA label is used as the reference annotation only for measuring cross-estimator agreement, rather than as verified chronological ground truth.
For each independent evaluator, we report the mean absolute age difference from the MiVOLO label, the percentages of predictions within $\pm5$ and $\pm10$ years, and the Pearson correlation coefficient.
We additionally construct an external consensus by taking the per-image median of the four independent age predictions.
\end{revblock}

\begin{revblock}
As shown in Table~\ref{tab:hffa_age_label_consistency}, the four independent evaluators show consistent agreement with the MiVOLO-derived labels despite their different architectures and training data. The external median consensus further improves all agreement measures, achieving a mean absolute difference of 3.47 years and a Pearson correlation of 0.945. Two non-numerical Phi-4 responses are excluded without manual imputation. A leave-one-evaluator-out analysis in Sec.~I-A of the supplementary material further shows that the consensus is not dominated by any single evaluator.
\end{revblock}

\subsubsection{Averaged Age Representations in the Age Codebook}

A critical component of HFFA is the \textbf{age codebook}, which is constructed by averaging the age embeddings assigned to each specific numerical age to obtain \rev{a shared age representation.}

\begin{revblock}
For each numerical age $a\in\{1,\ldots,85\}$, let $\mathcal{I}_a=\{j\mid a_j=a\}$ denote the set of valid training samples assigned to age $a$, where $a_j=\operatorname{round}(\tilde{a}_j)$.
The corresponding age-codebook entry is computed as the direct element-wise mean
\begin{equation}
    e_{\mathrm{code}}^{a}
    =
    \frac{1}{|\mathcal{I}_a|}
    \sum_{j\in\mathcal{I}_a}
    e_{\mathrm{age}}^{j}.
\end{equation}
The resulting codebook therefore provides one shared 384-dimensional age representation for each integer age from 1 to 85.
\end{revblock}

\begin{revblock}
For this analysis, the official FairFace \texttt{race\_4} model~\cite{Karkkainen_2021_FairFace,FairFace_Repository} is applied retrospectively to the HFFA training set.
For reporting, its four predefined output categories are denoted as Groups 1--4 following the fixed output-index order used throughout our group-wise analyses.
For each group, we independently average the age embeddings at each numerical age, compute the cosine similarity between the group-specific and overall codebook entries at the same age, and report the arithmetic mean of the per-age similarities.
All four groups contain valid samples at every integer age from 1 to 85; therefore, no age bin is omitted or interpolated in this analysis.
\end{revblock}

We compare group-specific age codebooks with the overall codebook \rev{to characterize representation-level consistency across the four FairFace-assigned output groups.}

As shown in \tabref{tab:codebook}, all group-specific age codebooks have cosine similarities above 0.97 with the overall codebook. \rev{These results indicate that the averaged age representations remain similar across Groups 1--4 at the representation level. Editing-outcome consistency across the same four groups is evaluated separately using group-wise age accuracy and identity preservation.}

\begin{table}[h]
\normalsize
    \centering
    \vspace{-5pt}
    \caption{Similarity between group-specific and overall age codebooks}
    \renewcommand{\arraystretch}{1.3}
    \centering
        \begin{tabular}{@{}c|cccc@{}}
\toprule
Group                   & Group 1 & Group 2 & Group 3 & Group 4 \\ \midrule
Similarity $\uparrow$   & 0.998   & 0.976   & 0.994   & 0.989   \\ \bottomrule
\end{tabular}
\vspace{-15pt}

        \label{tab:codebook}
\end{table}

\subsection{Experimental Setup}

\subsubsection{Training and Model Selection}

TimeMachine is trained on the SD1.5 base model in two stages \rev{using BF16 precision at a training resolution of $512\times512$}.
\rev{Both stages use Adam with $(\beta_1,\beta_2)=(0.9,0.999)$, a weight decay of $10^{-4}$, and a constant learning-rate schedule.}
\rev{The original SD1.5 UNet, VAE, and text encoder weights remain frozen throughout training.}
Stage I optimizes \rev{the condition-projection and Multi-CA modules} without ACG \rev{for 700,000 steps with a learning rate of $1\times10^{-4}$}.
Stage II \rev{continues to optimize these modules jointly with ACG for another 5,000 steps with a learning rate of $5\times10^{-6}$ and $\lambda_{\mathrm{ACG}}=0.05$}.
Both stages use eight NVIDIA H20 GPUs with a batch size of 8 per GPU.
\rev{All four Multi-CA branch coefficients are fixed to 1.0 during training.}

\begin{revblock}
The fixed 100-image HFFA validation subset is excluded from both training stages and age-codebook construction and is used only for periodic qualitative assessment and checkpoint selection; the remaining processed HFFA images constitute the training set.
During Stage I, we monitor age controllability, identity preservation, and visual quality and retain the 700,000-step checkpoint after these behaviors become stable.
During Stage II, we additionally monitor the convergence of the ACG age-embedding loss while confirming that the diffusion denoising loss and qualitative editing behavior remain stable.
No images or evaluation metrics from the final public test benchmarks are used for checkpoint selection or hyperparameter adjustment.
\end{revblock}

\begin{table*}[htb]
    \small
    \centering
    \caption{\rev{Quantitative comparison on the 500-image CelebA-HQ test set.} Age MAE is evaluated using MiVOLO, identity similarity is evaluated using AdaFace, and FaceQ and Blur are obtained from FACE++. The best result is shown in \textbf{bold}, and the second best is \underline{underlined}.}
    \renewcommand{\arraystretch}{1.3}
    \begin{tabular}{@{}c|cccccccc|c|cc@{}}
        \toprule
        \multirow{2}{*}{\diagbox{Method}{Metric}}
        & \multicolumn{8}{c|}{Age MAE $\downarrow$}
        & \multirow{2}{*}{AdaFace $\uparrow$}
        & \multicolumn{2}{c}{Image Quality} \\
        & 10 & 20 & 30 & 40 & 50 & 60 & 70 & Average
        & & FaceQ $\uparrow$ & Blur $\downarrow$ \\
        \midrule
        LATS
        & 3.142 & 3.611 & \underline{4.306} & 5.417
        & 6.012 & 6.084 & 15.315 & 6.270
        & 0.474 & 68.532 & 0.883 \\
        CUSP
        & 19.661 & 3.060 & 6.317 & 7.356
        & \underline{4.897} & 4.505 & 39.717 & 12.216
        & 0.576 & 67.396 & 0.993 \\
        SAM
        & 5.406 & 4.278 & 4.570 & \underline{4.850}
        & 5.315 & 5.088 & 5.908 & 5.059
        & 0.413 & 70.131 & 1.699 \\
        IP-Adapter
        & 5.308 & 2.665 & 6.700 & 7.044
        & 6.328 & 6.872 & 9.402 & 6.331
        & 0.520 & 73.653 & \underline{0.536} \\
        PhotoMaker
        & 18.746 & 10.714 & 12.177 & 15.870
        & 21.590 & 29.082 & 37.655 & 20.833
        & 0.352 & \underline{74.581} & 0.995 \\
        FADING
        & \underline{2.316} & \underline{2.232} & 5.819 & 6.482
        & 6.671 & \underline{4.001} & \underline{5.050}
        & \underline{4.653} & \underline{0.668} & 66.723 & 0.668 \\
        Ours
        & \textbf{1.904} & \textbf{1.785} & \textbf{3.365}
        & \textbf{3.971} & \textbf{3.655} & \textbf{3.072}
        & \textbf{3.814} & \textbf{3.081}
        & \textbf{0.682} & \textbf{75.228} & \textbf{0.479} \\
        \bottomrule
    \end{tabular}
    \label{tab:main_500}
\end{table*}

\subsubsection{Testing datasets}
\leavevmode
HFFA supplies the training and \rev{fixed validation} data, while CelebA-HQ, MORPH, AgeDB, FFHQ, and CACD serve as \rev{external} quantitative-evaluation benchmarks.

CelebA-HQ~\cite{karras2017progressive} serves as the primary test benchmark.
\begin{revblock}
We select 500 source images using stratified sampling over MiVOLO-estimated age and gender and edit each image toward seven target ages: 10, 20, 30, 40, 50, 60, and 70.
MiVOLO-estimated age and gender define the sampling strata independently of generated outputs and evaluation scores.
The same 500 images and seven target ages are used for all compared methods and evaluators.
\end{revblock}
We evaluate LATS, CUSP, SAM, IP-Adapter, PhotoMaker, FADING, and TimeMachine \rev{under the same protocol, producing 3,500 generated images per method and 24,500 generated images in total.}
\rev{All five age evaluators are applied to the same fixed set of generated images.}

We additionally evaluate cross-dataset generalization on MORPH \cite{ricanek2006morph}, AgeDB \cite{AGEDB8014984}, FFHQ \cite{karras2019style}, and CACD \cite{chen2014cross}.
\begin{revblock}
For each dataset, we select 500 source images using the same age-and-gender stratification protocol and apply the same seven target ages, yielding 2,000 cross-dataset source images in total.
To evaluate cross-group robustness, we additionally sample 400 CelebA-HQ images that do not overlap with the 500-image main test set, with 100 images from each of the four FairFace-assigned output groups, denoted as Groups 1--4.
Within each group, the samples are stratified over age and gender, and SAM, FADING, and TimeMachine are evaluated using the same seven target ages.

We screen all 2,900 final quantitative-evaluation images against the complete HFFA collection using SHA-256 hashing, perceptual hashing, and Antelopev2 face-embedding retrieval with AdaFace rescoring and final manual verification; after replacing confirmed matches and repeating the audit, no confirmed image-level or identity-level overlap remains.
\end{revblock}

\subsubsection{\rev{TimeMachine Inference Configuration}}

\rev{
All main quantitative and qualitative TimeMachine results are generated using the Stable Diffusion v1.5 img2img pipeline (\texttt{StableDiffusionImg2ImgPipeline}) with \texttt{sd-vae-ft-mse}.
We use 50 DDIM steps, a CFG scale of 2.5, an img2img strength of 1.0, \texttt{padding}=75, a fixed seed, and an output resolution of $512\times512$.
The fixed condition settings are $\lambda_{\mathrm{ID}}=0.75$, $\lambda_{\mathrm{age}}=1.75$, $\lambda_{\mathrm{age\text{-}text}}=1.75$, and a LoRA scale of 1.0.
These settings remain unchanged across datasets, source images, demographic groups, and target ages.
}

\subsubsection{\rev{Matched Computational Profile}}

\begin{revblock}
We profile the SD v1.5 reference and TimeMachine with the same SD v1.5 backbone, FP16 precision, $512\times512$ resolution, batch size of 1, and 50 DDIM steps; latency is averaged over 100 fixed source--target pairs on an RTX 3090, including condition construction and CPU-based Antelopev2 extraction but excluding model loading and disk I/O.
\end{revblock}

\begin{table}[h]
\vspace{-10pt}
\centering
\small
\caption{\rev{Matched computational profile.}}
\label{tab:computational_profile}
\setlength{\tabcolsep}{3.5pt}
\resizebox{\linewidth}{!}{%
\begin{tabular}{@{}lccc@{}}
\toprule
Model & UNet FLOPs/step $\downarrow$ & Peak memory $\downarrow$ & Latency/image $\downarrow$ \\
\midrule
SD1.5 reference & 1606.78 G & 3.071 GiB & 3.1283 s \\
TimeMachine & 1686.89 G & 3.311 GiB & 4.0633 s \\
\bottomrule
\end{tabular}%
}

\end{table}

\begin{revblock}
As shown in \tabref{tab:computational_profile}, the TimeMachine conditioning modules add 4.99\% UNet FLOPs per step and 29.89\% latency relative to the matched SD1.5 reference. The SD1.5 row isolates the overhead introduced by these modules, while training-only ACG leaves inference cost unchanged.
\end{revblock}

\subsubsection{Evaluation Metrics}
\leavevmode

We evaluate age accuracy, identity preservation, and image quality.
For age accuracy, Age MAE is computed as the absolute difference between an evaluator's predicted age and the specified target age.  
\begin{revblock}
For each target age, the MAE is averaged over the 500 source images; the overall MAE is averaged over all 3,500 generated images.
In addition to MiVOLO~\cite{kuprashevich2023mivolo}, we use four complementary age estimators exclusively for evaluation.
MIVIA SE-ResNet50~\cite{Greco_2022_MIVIA} is a CNN-based regression estimator whose scalar output is used as the predicted age.
Faceptor-Full~\cite{Qin_2024_Faceptor} is a Transformer-based face-perception model whose age prediction is decoded as the expectation of its output distribution.
For zero-shot VLM-based evaluation, we use Phi-4 Multimodal~\cite{Microsoft_2025_Phi4} and Qwen3-VL-8B~\cite{Bai_2025_Qwen3VL} under the common protocol of AgeMoVLE~\cite{Chiu_2026_AgeMoVLE}.
Phi-4 achieved the lowest overall MAE among the five individual frozen VLM experts evaluated in AgeMoVLE, while Qwen3-VL-8B provides a complementary VLM family with dynamic-resolution visual tokenization.
Both VLMs use deterministic decoding and the fixed prompt ``You are a professional face analyst. Estimate the person's age and reply ONLY with a number,'' and the first valid numerical value in each response is taken as the predicted age.
All evaluators use their official checkpoints and inference settings.
\end{revblock}

Identity preservation is measured by the cosine similarity between AdaFace \cite{kim2022adaface} embeddings extracted from the source and generated images.
\begin{revblock}
Before automatic evaluation, outputs from all methods are resized to a common resolution of $512\times512$, after which each metric applies its official preprocessing protocol.
\end{revblock}

\subsection{Comparison with Other Age Datasets} \label{subsec:com}

To contextualize HFFA, we compare it with several commonly used age-related face datasets: MORPH Album II (MORPH) \cite{ricanek2006morph}, AgeDB \cite{AGEDB8014984}, CACD \cite{chen2014cross}, and FFHQ-Aging \cite{or2020lifespan,karras2019style}.
These resources differ in image resolution, scale, age coverage, and annotation granularity.

\begin{table}[h]
\normalsize
    \centering
\caption{Comparison of different age datasets}
    \renewcommand{\arraystretch}{1.3}
    \begin{tabular}{@{}c|ccc@{}}
\toprule
Dataset    & Resolution & Size & Age Range \\ \midrule
HFFA       & $1024\times1024$ & {Approx. 1M} & 1--85 \\
MORPH      & $180\times180$   & {55,134} & {16--77} \\
AgeDB      & Variable          & {16,488} & 1--101 \\
CACD       & $250\times250$   & {163,446} & 14--62 \\
FFHQ-Aging & $1024\times1024$ & {70,000} & {0--2 to 70+} \\ \bottomrule
\end{tabular}
\label{tab:dif-dataset}
\end{table}

As shown in \tabref{tab:dif-dataset}, HFFA combines a million-scale collection standardized to $1024\times1024$ resolution with \rev{model-estimated ages} covering 1--85 years.
MORPH and AgeDB provide smaller but valuable datasets with year-level age annotations, CACD provides a larger in-the-wild celebrity collection, and FFHQ-Aging augments 70,000 high-resolution FFHQ images with ten crowdsourced age groups ranging from 0--2 to 70+.
Together, these datasets provide complementary age coverage, image characteristics, and annotation granularity.

\subsection{Comparisons with State-of-the-art Methods}

\subsubsection{{Comparison Methods and Settings}}

We compare TimeMachine with the dedicated age-editing methods LATS~\cite{or2020lifespan}, CUSP~\cite{gomez2022custom}, SAM~\cite{alaluf2021only}, and FADING~\cite{chen2023face}, as well as the identity-preserving personalized-generation methods IP-Adapter-FaceID~\cite{ye2023ip} and PhotoMaker~\cite{li2024photomaker}.
\rev{All baselines use their public implementations and official checkpoints without retraining on HFFA or tuning on the test set; the same source identities and target ages are used with fixed settings for all samples.}

\begin{revblock}
LATS uses its official gender-specific models, with input gender estimated by the fixed Antelopev2 estimator and target outputs selected from its official age-anchor traversal.
CUSP uses the official FFHQ-RR model with the numerical DEX age condition, SAM directly receives the numerical target age, and FADING uses the Antelopev2-estimated source age together with the specified target age.

For IP-Adapter-FaceID, we use its official SD1.5 FaceID checkpoint with Realistic Vision V4.0 as the base model (scale 1.0, 30 DDIM steps, CFG 7.5, and $512\times512$ resolution).
The fixed prompt is ``a portrait photo of a [target age] year old [man/woman], photorealistic, high quality, detailed face.''

For PhotoMaker, we use its official V1 checkpoint with RealVisXL V3.0 as the base model and one reference image (50 Euler steps, CFG 5.0, merge step 10, and $1024\times1024$ resolution).
The fixed prompt is ``a photo of a [man/woman] img at the age of [target age], photorealistic, high quality.''
\end{revblock}

\subsubsection{Qualitative Comparison}

We first examine facial age editing at the extreme target ages of 10 and 80 in \figref{fig:demo_10_and_80}.
In the representative examples shown, LATS~\cite{or2020lifespan} and CUSP~\cite{gomez2022custom} introduce age-related changes but may lose identity details or produce over-smoothed textures.
SAM~\cite{alaluf2021only} and IP-Adapter~\cite{ye2023ip} retain recognizable aspects of the input identity, although their target-age changes are less pronounced in some examples, particularly at age 10.
PhotoMaker produces relatively limited age shifts in the displayed examples, whereas FADING~\cite{chen2023face} introduces more visible age cues but occasionally exhibits local structural distortions.
TimeMachine produces clearer changes toward both extreme targets while retaining recognizable identity-related facial structure.

\rev{These examples illustrate the qualitative behavior of the compared methods; aggregate age accuracy, identity preservation, and image quality are evaluated in the following quantitative experiments.}

We further compare TimeMachine with SAM and FADING over a progressive sequence of target ages from 10 to 80 in \figref{fig:demo_10_to_80}.
In the shown sequences, SAM changes the input more conservatively across age intervals, while FADING introduces pronounced aging cues but occasionally shows local artifacts or structural inconsistencies.
TimeMachine exhibits gradual changes in age-related facial texture, hair appearance, and facial geometry while retaining subject-specific characteristics across the sequence.

\subsubsection{Quantitative Evaluation}

As shown in \tabref{tab:main_500}, TimeMachine achieves the lowest MiVOLO Age MAE at every target age \rev{on the 500-image CelebA-HQ test set, with an overall average of 3.081}. \rev{The consistent advantage across all seven target ages demonstrates the effectiveness of TimeMachine for fine-grained age control over the complete evaluated age range.} It also obtains the highest AdaFace identity similarity, \rev{indicating that accurate target-age transformation and identity preservation can be achieved simultaneously}. \rev{Moreover, the highest Face Quality score and the lowest Image Blur score show that these improvements are obtained without reducing facial quality or image sharpness. This visual-quality advantage may also benefit from the standardized high-quality preprocessing used in constructing HFFA.}

\begin{figure*}[!t]
\centering
\begin{minipage}[t]{0.49\textwidth}
\centering
\includegraphics[width=\linewidth]{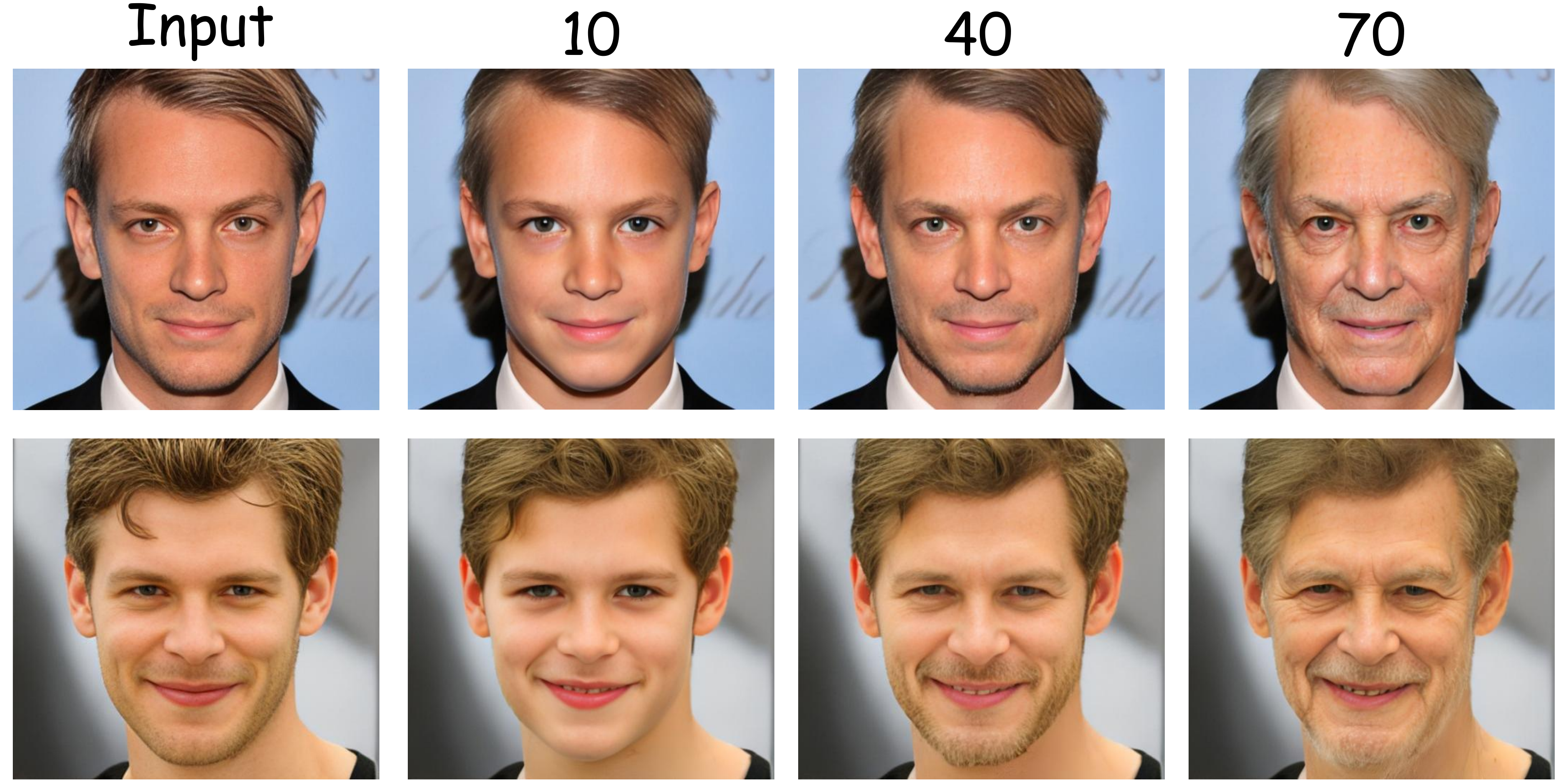}\\[-0.25em]
\textbf{(a) Group 1}
\end{minipage}
\hfill
\begin{minipage}[t]{0.49\textwidth}
\centering
\includegraphics[width=\linewidth]{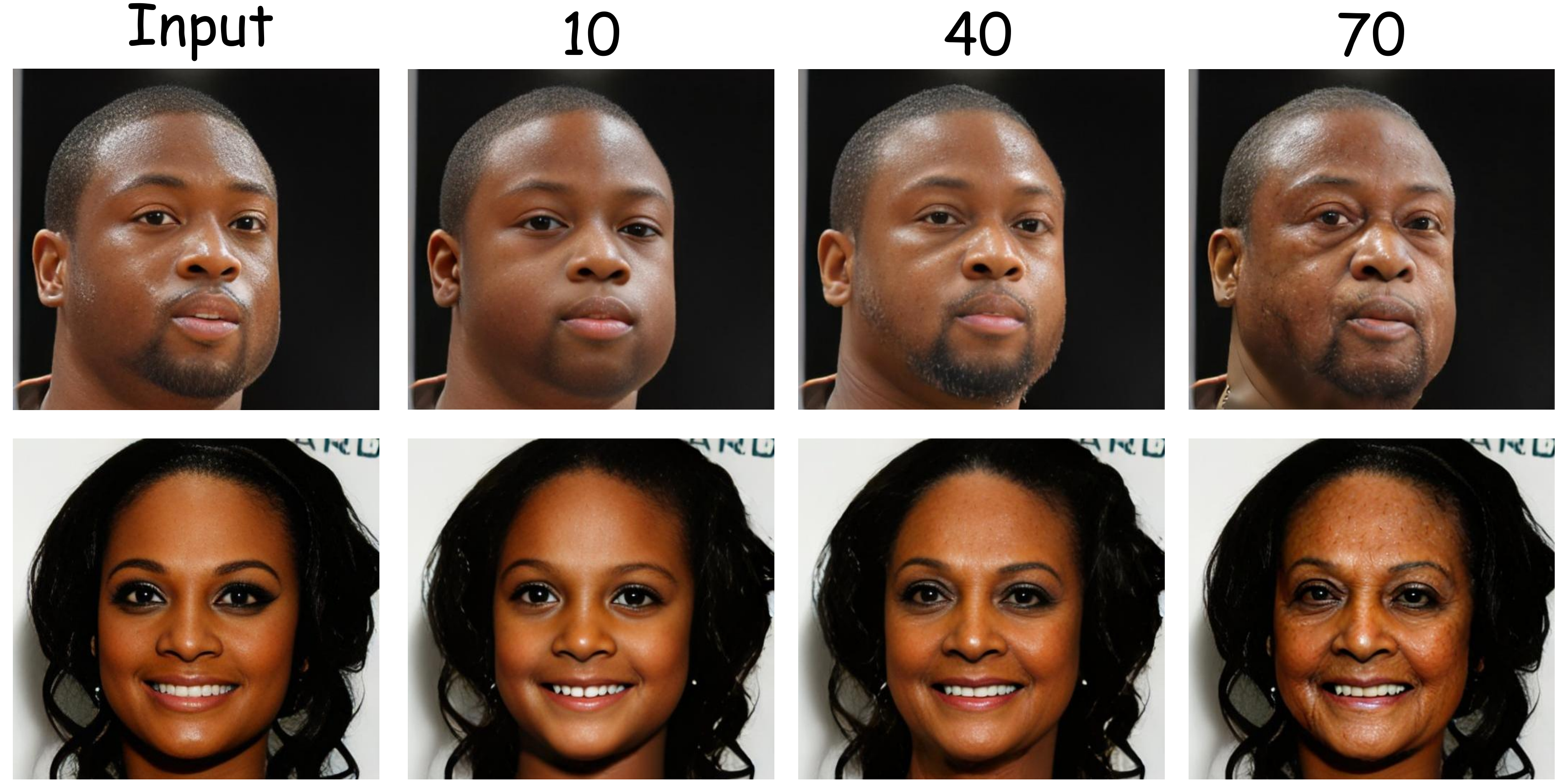}\\[-0.25em]
\textbf{(b) Group 2}
\end{minipage}

\vspace{0.4em}

\begin{minipage}[t]{0.49\textwidth}
\centering
\includegraphics[width=\linewidth]{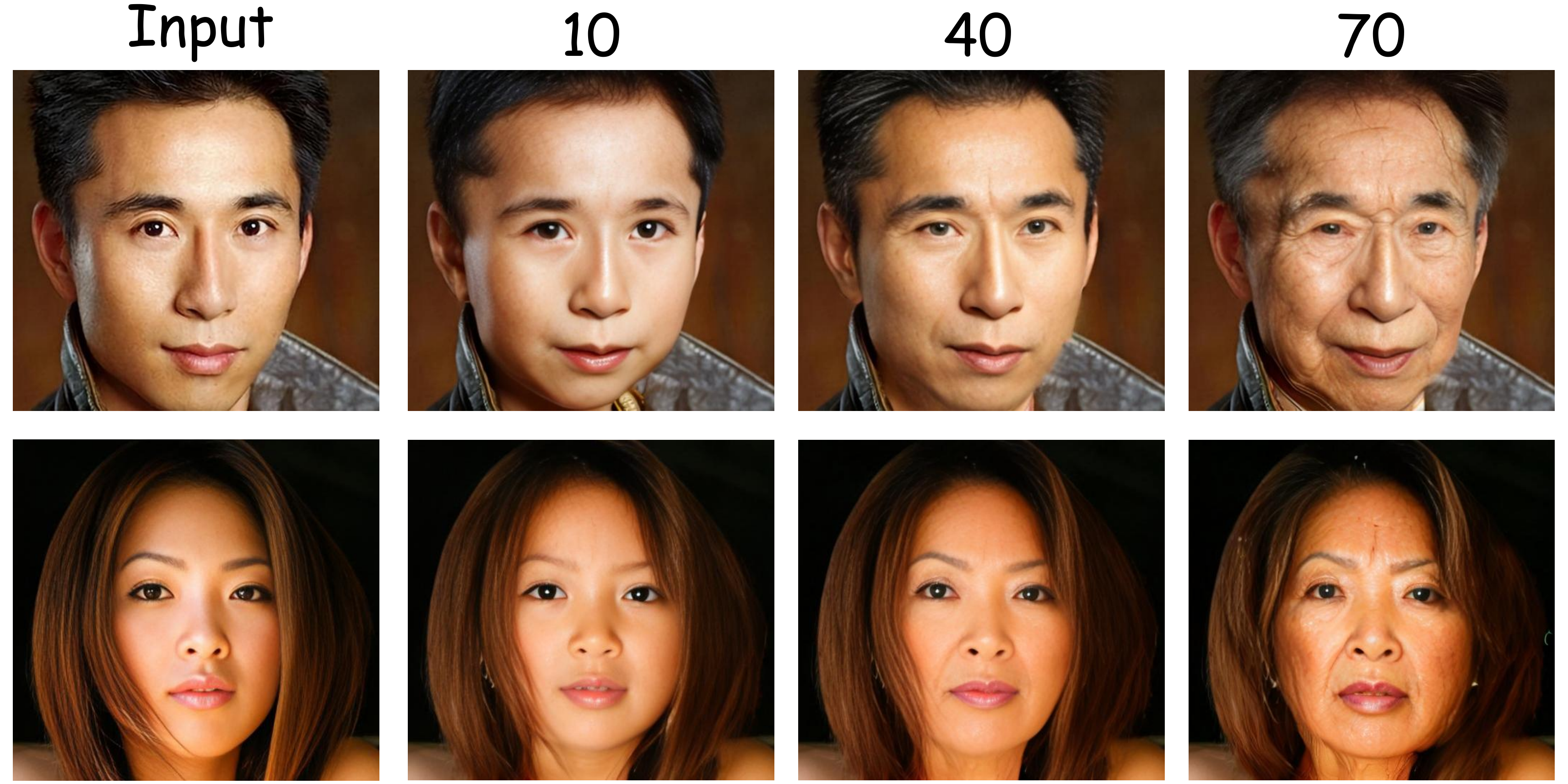}\\[-0.25em]
\textbf{(c) Group 3}
\end{minipage}
\hfill
\begin{minipage}[t]{0.49\textwidth}
\centering
\includegraphics[width=\linewidth]{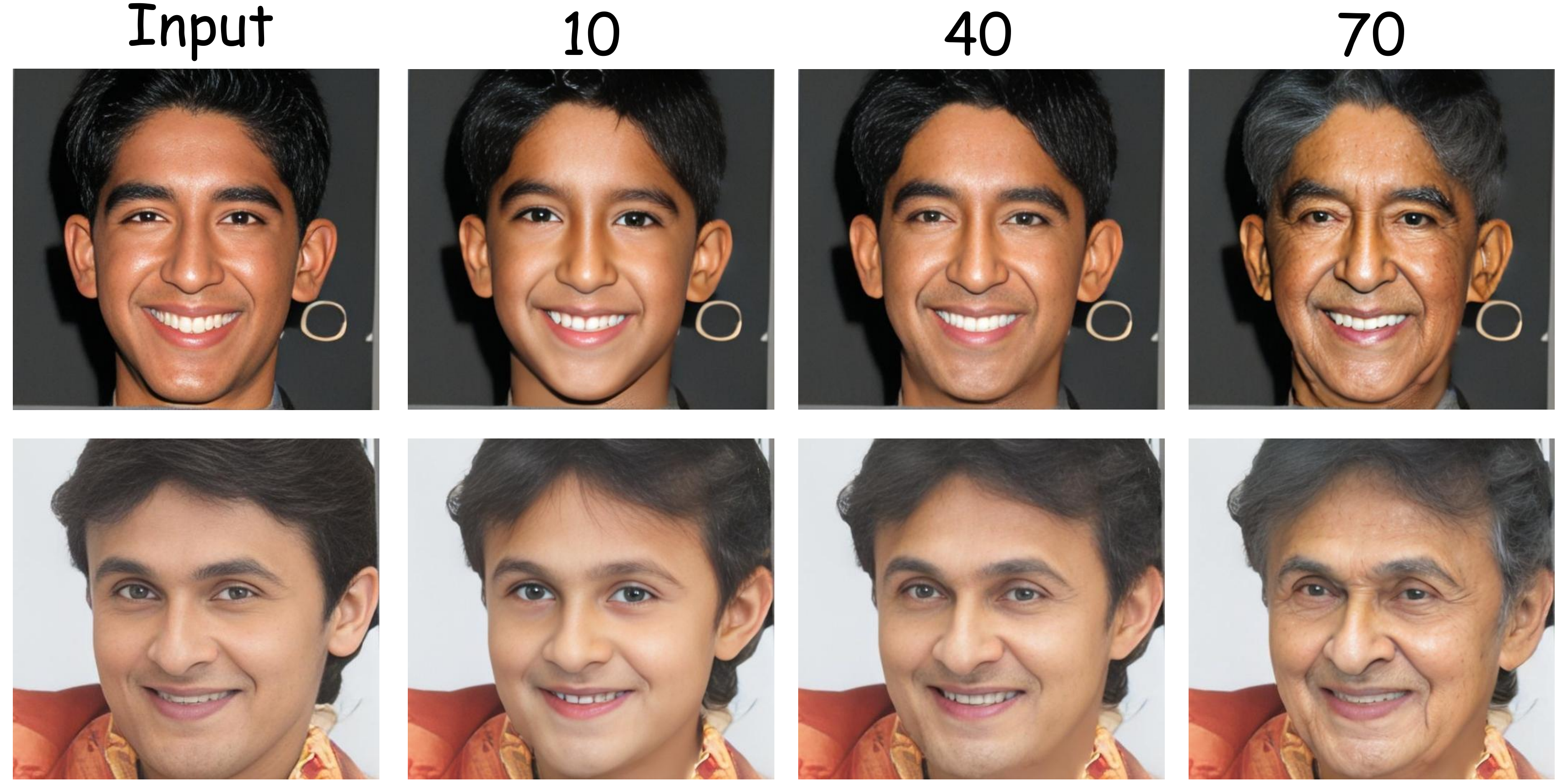}\\[-0.25em]
\textbf{(d) Group 4}
\end{minipage}
\caption{\rev{Representative qualitative results across FairFace-assigned Groups 1--4. Each example shows the input and edited results at target ages of 10, 40, and 70.}}
\label{fig:demographic_qualitative}
\end{figure*}

\begin{table}[!t]
\centering
\small
\caption{\rev{Average Age MAE ($\downarrow$) on the 500-image CelebA-HQ test set under five age evaluators. The best result is shown in \textbf{bold}, and the second best is \underline{underlined}.}}
\renewcommand{\arraystretch}{1.2}
\setlength{\tabcolsep}{2.2pt}
\begin{tabular*}{\columnwidth}{@{\extracolsep{\fill}}c|ccccc@{}}
\toprule
Method & MiVOLO & MIVIA & Faceptor & Phi-4 & Qwen3-VL \\ \midrule
LATS       & 6.270  & 10.592 & 11.860 & 7.267  & 7.381 \\
CUSP       & 12.216 & 12.918 & 14.246 & 13.099 & 13.849 \\
SAM        & 5.059  & 8.151  & 10.884 & 6.475  & \underline{6.607} \\
IP-Adapter & 6.331  & 8.824  & \underline{9.418} & \underline{6.258} & 6.785 \\
PhotoMaker & 20.833 & 19.627 & 18.656 & 20.124 & 20.265 \\
FADING     & \underline{4.653} & \underline{7.900} & 10.268 & 6.765 & 7.284 \\
Ours       & \textbf{3.081} & \textbf{5.565} & \textbf{8.847} & \textbf{4.135} & \textbf{4.969} \\ \bottomrule
\end{tabular*}
\label{tab:cross_evaluator_500}
\end{table}

\begin{revblock}
As shown in \tabref{tab:cross_evaluator_500}, TimeMachine achieves the lowest average MAE under all five age estimators.
Although the absolute errors and the relative ordering among the baselines vary across evaluators, TimeMachine consistently ranks first under the CNN-based regression model, the Transformer-based face-perception model, and both zero-shot VLM evaluators.
The complete per-target-age results for MIVIA, Faceptor, Phi-4, and Qwen3-VL are reported in Sec.~II of the supplementary material.
\end{revblock}
\begin{table}[!ht]
\centering
\scriptsize
\setlength{\tabcolsep}{1.2pt}
\caption{\rev{Cross-dataset results on 500 images per dataset. Best in \textbf{bold}.}}
\label{tab:more-test}
\renewcommand{\arraystretch}{1.05}
\resizebox{\columnwidth}{!}{%
\begin{tabular}{@{}l|ccc|ccc|ccc@{}}
\toprule
\multirow{2}{*}{Dataset}
& \multicolumn{3}{c|}{MiVOLO MAE $\downarrow$}
& \multicolumn{3}{c|}{MIVIA MAE $\downarrow$}
& \multicolumn{3}{c}{AdaFace Sim. $\uparrow$} \\
& SAM & FADING & Ours
& SAM & FADING & Ours
& SAM & FADING & Ours \\
\midrule
MORPH
& 4.154 & 6.958 & \textbf{4.101}
& 8.633 & 8.506 & \textbf{7.030}
& 0.355 & 0.617 & \textbf{0.627} \\
AgeDB
& 5.188 & 6.371 & \textbf{3.499}
& 8.567 & 8.429 & \textbf{7.109}
& 0.423 & 0.658 & \textbf{0.692} \\
FFHQ
& 4.677 & 7.903 & \textbf{4.426}
& 8.201 & 8.222 & \textbf{6.982}
& 0.385 & 0.596 & \textbf{0.630} \\
CACD
& 4.705 & 6.066 & \textbf{3.383}
& 7.410 & 11.855 & \textbf{7.255}
& 0.410 & 0.584 & \textbf{0.627} \\
\midrule
Overall
& 4.681 & 6.825 & \textbf{3.852}
& 8.203 & 9.253 & \textbf{7.094}
& 0.393 & 0.614 & \textbf{0.644} \\
\bottomrule
\end{tabular}
}
\end{table}

\rev{As shown in \tabref{tab:more-test}, TimeMachine ranks first in both age accuracy metrics and identity similarity on MORPH, AgeDB, FFHQ, and CACD.}
The consistent gains across datasets with different age distributions, demographic compositions, and image characteristics indicate that the observed age--identity balance generalizes beyond the primary CelebA-HQ evaluation.

\subsubsection{{Demographic-Group Evaluation}}

\begin{revblock}
After excluding the 500 CelebA-HQ images in the main evaluation, we independently sample another 400 non-overlapping images from the remaining data.
The official FairFace \texttt{race\_4} model~\cite{Karkkainen_2021_FairFace,FairFace_Repository} assigns each candidate to one of four predefined output categories, which are consistently denoted as Groups 1--4 according to the fixed output-index order.
Each group contains 100 source images selected through stratified random sampling over age and gender, with a balanced gender distribution.
Following the main evaluation protocol, SAM, FADING, and TimeMachine edit every image toward target ages of 10, 20, 30, 40, 50, 60, and 70, producing 8,400 images in total; no generation or metric-evaluation failure occurs.
\end{revblock}

\begin{table}[h]
\centering

\scriptsize
\setlength{\tabcolsep}{1.2pt}
\caption{\rev{Results across FairFace-assigned Groups 1--4, with 100 images per group. Best in \textbf{bold}.}}
\label{tab:demographic_evaluation}
\renewcommand{\arraystretch}{1.05}
\resizebox{\columnwidth}{!}{%
\begin{tabular}{@{}l|ccc|ccc|ccc@{}}
\toprule
\multirow{2}{*}{Group}
& \multicolumn{3}{c|}{MiVOLO MAE $\downarrow$}
& \multicolumn{3}{c|}{MIVIA MAE $\downarrow$}
& \multicolumn{3}{c}{AdaFace Sim. $\uparrow$} \\
& SAM & FADING & Ours
& SAM & FADING & Ours
& SAM & FADING & Ours \\
\midrule
Group 1
& 5.461 & 5.171 & \textbf{3.096}
& 8.407 & 8.296 & \textbf{5.014}
& 0.399 & 0.605 & \textbf{0.689} \\
Group 2
& 5.632 & 3.698 & \textbf{3.442}
& 8.362 & 7.598 & \textbf{7.417}
& 0.391 & 0.653 & \textbf{0.705} \\
Group 3
& 4.450 & 4.405 & \textbf{3.414}
& 7.386 & 7.601 & \textbf{6.131}
& 0.384 & 0.634 & \textbf{0.671} \\
Group 4
& 4.763 & 3.298 & \textbf{3.156}
& 8.361 & 7.096 & \textbf{6.143}
& 0.429 & 0.678 & \textbf{0.700} \\
\midrule
Overall
& 5.077 & 4.143 & \textbf{3.277}
& 8.129 & 7.648 & \textbf{6.176}
& 0.401 & 0.643 & \textbf{0.691} \\
\bottomrule
\end{tabular}
}

\end{table}

\begin{figure*}[!t]

\centering
\begin{minipage}[t]{0.49\textwidth}
\centering
\includegraphics[width=\linewidth]{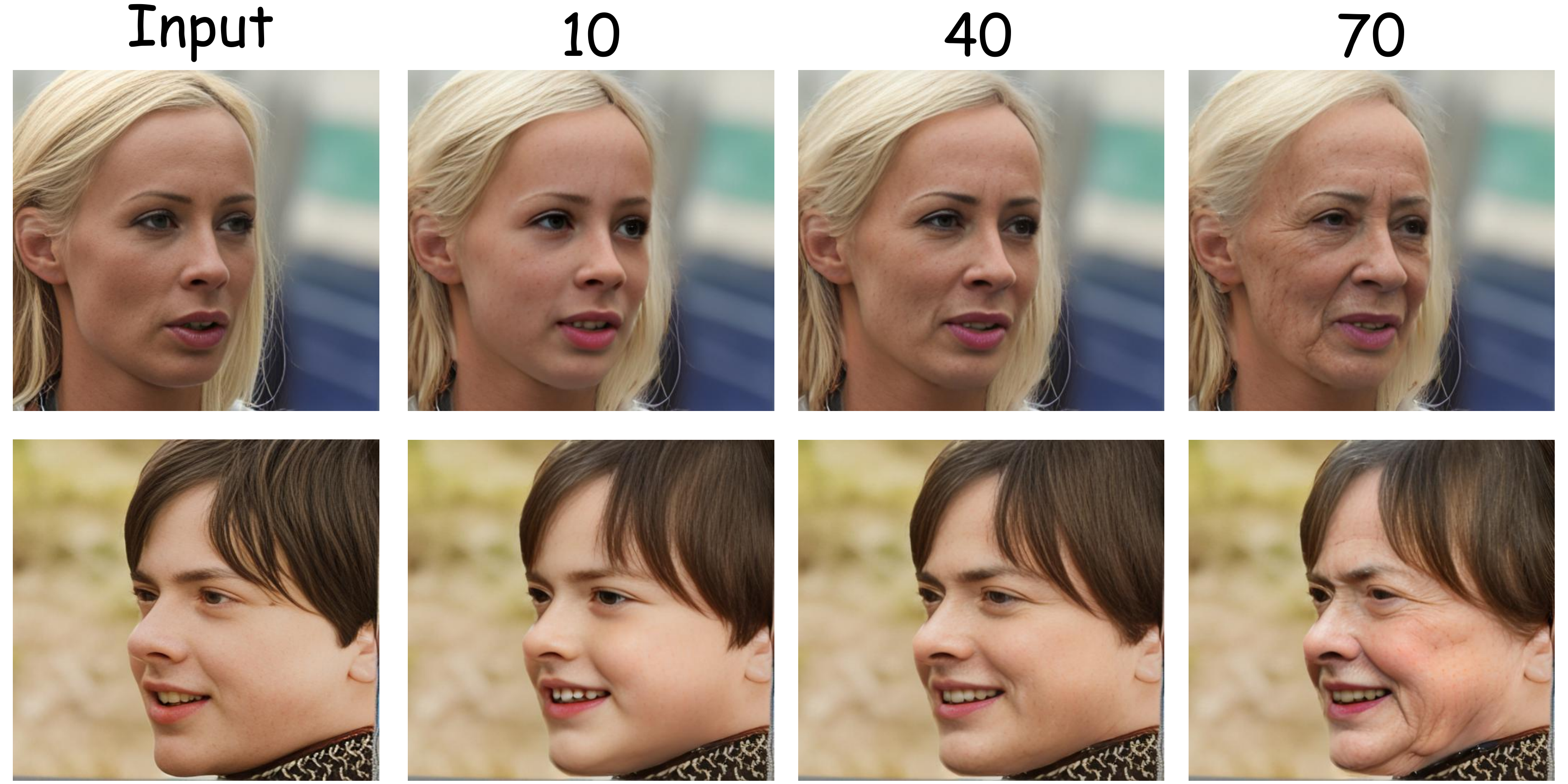}\\[-0.25em]
\textbf{(a) Profile views}
\end{minipage}
\hfill
\begin{minipage}[t]{0.49\textwidth}
\centering
\includegraphics[width=\linewidth]{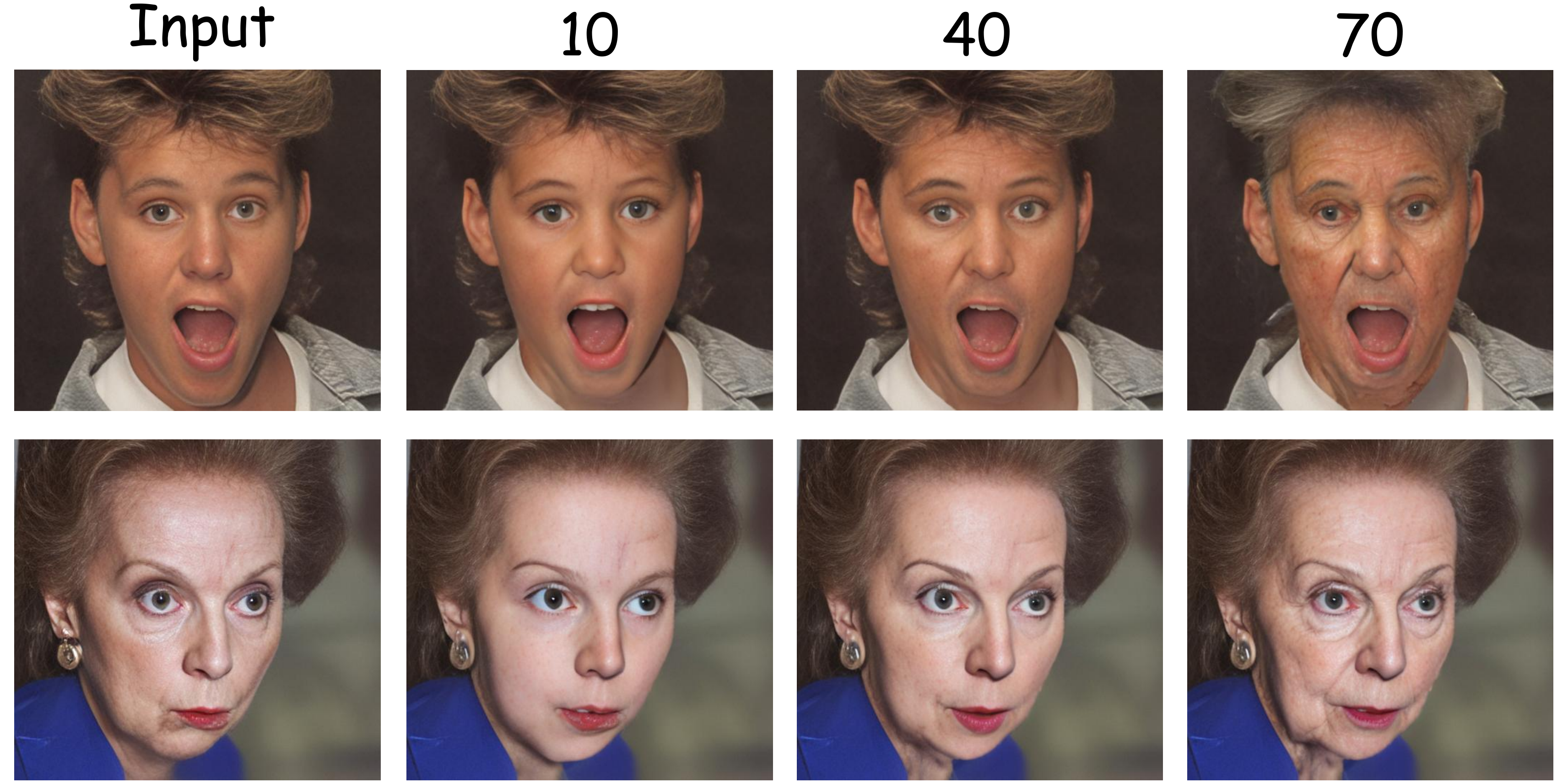}\\[-0.25em]
\textbf{(b) Exaggerated expressions}
\end{minipage}

\vspace{0.4em}

\begin{minipage}[t]{0.49\textwidth}
\centering
\includegraphics[width=\linewidth]{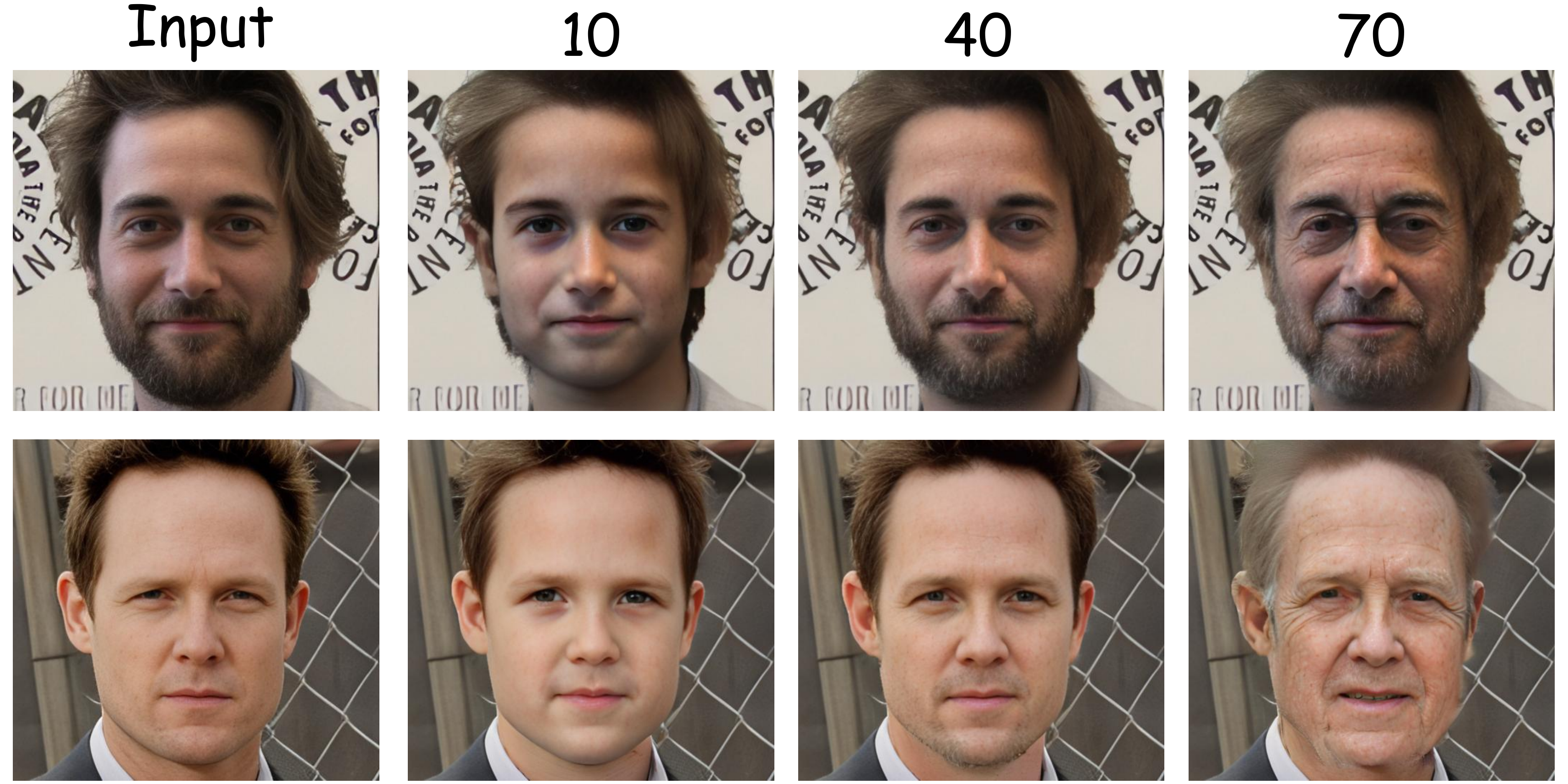}\\[-0.25em]
\textbf{(c) Complex backgrounds}
\end{minipage}
\hfill
\begin{minipage}[t]{0.49\textwidth}
\centering
\includegraphics[width=\linewidth]{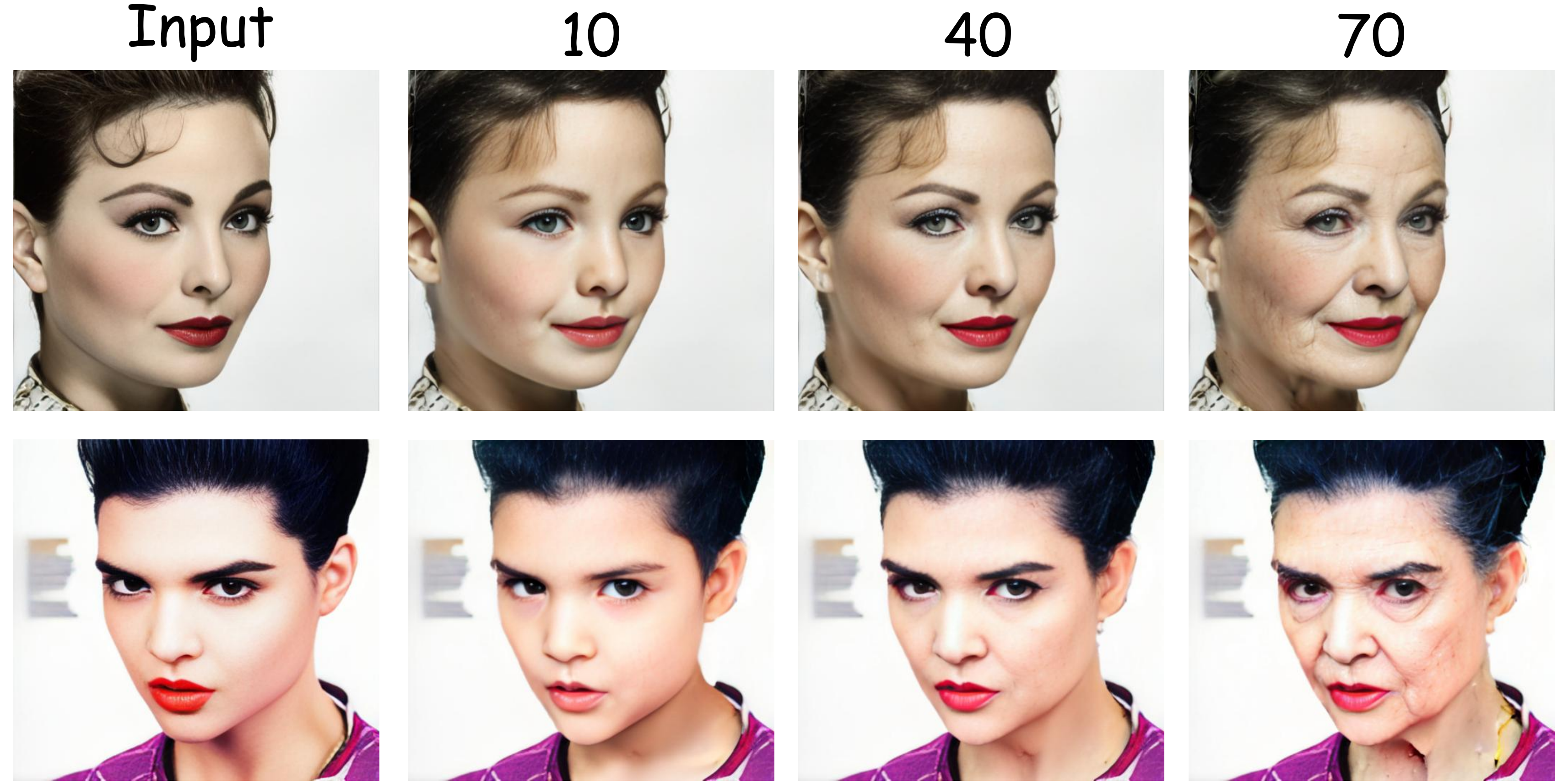}\\[-0.25em]
\textbf{(d) Makeup}
\end{minipage}

\vspace{0.4em}

\begin{minipage}[t]{0.49\textwidth}
\centering
\includegraphics[width=\linewidth]{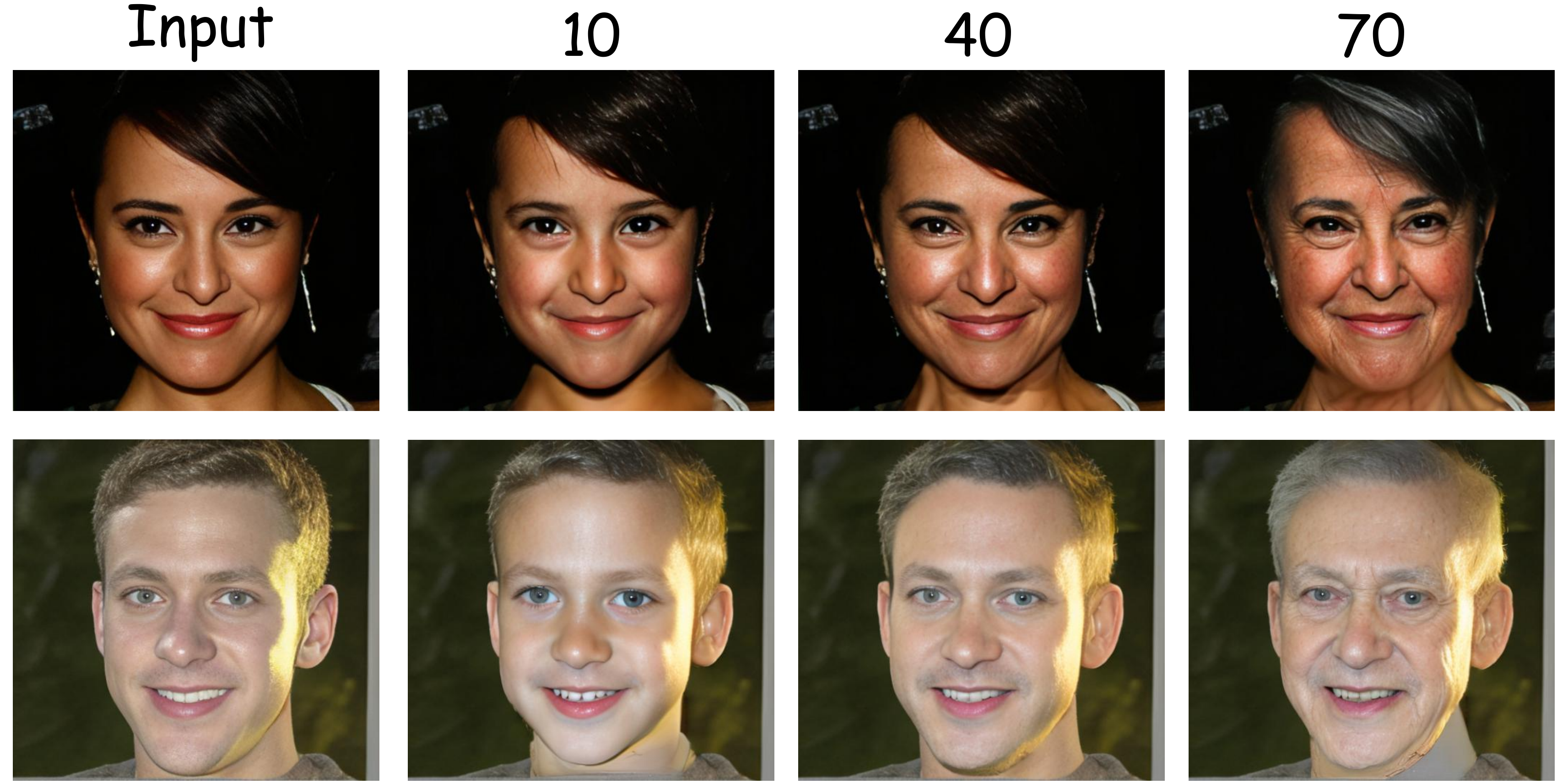}\\[-0.25em]
\textbf{(e) Low-light or low-quality inputs}
\end{minipage}
\hfill
\begin{minipage}[t]{0.49\textwidth}
\centering
\includegraphics[width=\linewidth]{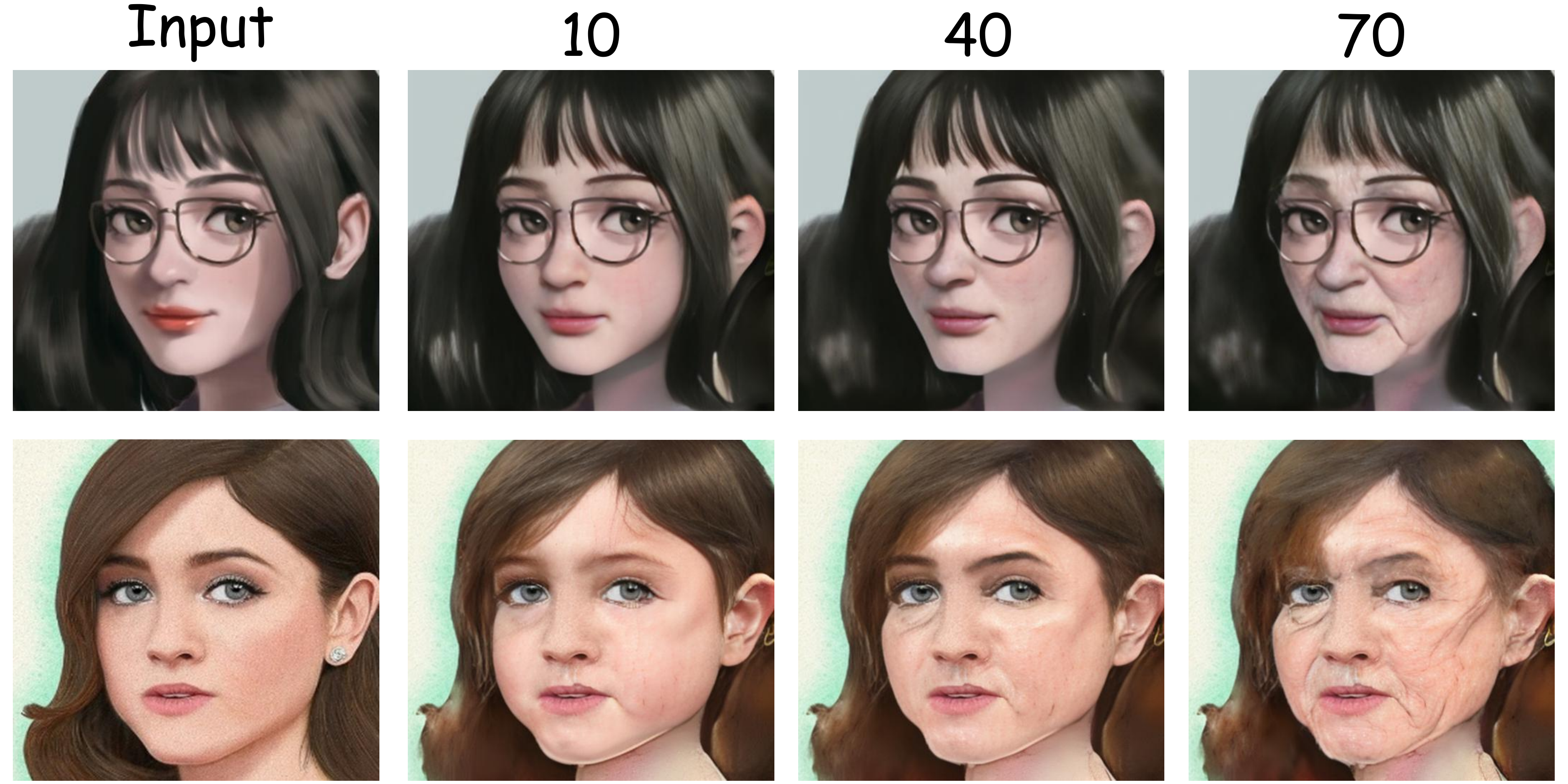}\\[-0.25em]
\textbf{(f) Artistic styles}
\end{minipage}
\caption{\rev{Representative qualitative results under diverse and challenging single-subject input conditions. Each input is edited toward target ages of 10, 40, 70.}}
\label{fig:diverse_input_conditions}
\end{figure*}

\begin{revblock}
As shown in \tabref{tab:demographic_evaluation}, TimeMachine ranks first in both age metrics and identity similarity across Groups 1--4. \rev{Although the absolute age errors vary across groups, the comparative ranking remains consistent under both MiVOLO and the independent MIVIA estimator, and TimeMachine obtains the highest AdaFace similarity in every group. This agreement across two age estimators and the identity metric indicates that the comparative advantage is not concentrated in a single model-assigned group or determined by one age evaluator.} The representative examples in \figref{fig:demographic_qualitative} further show progressive age-related changes while retaining the principal identity structure and input-specific facial characteristics. \rev{We interpret these results as evidence of cross-group comparative consistency within the four FairFace-assigned output groups and the present evaluation protocol.}
\end{revblock}

\begin{figure}[!ht]

\centering
\includegraphics[width=0.98\linewidth]{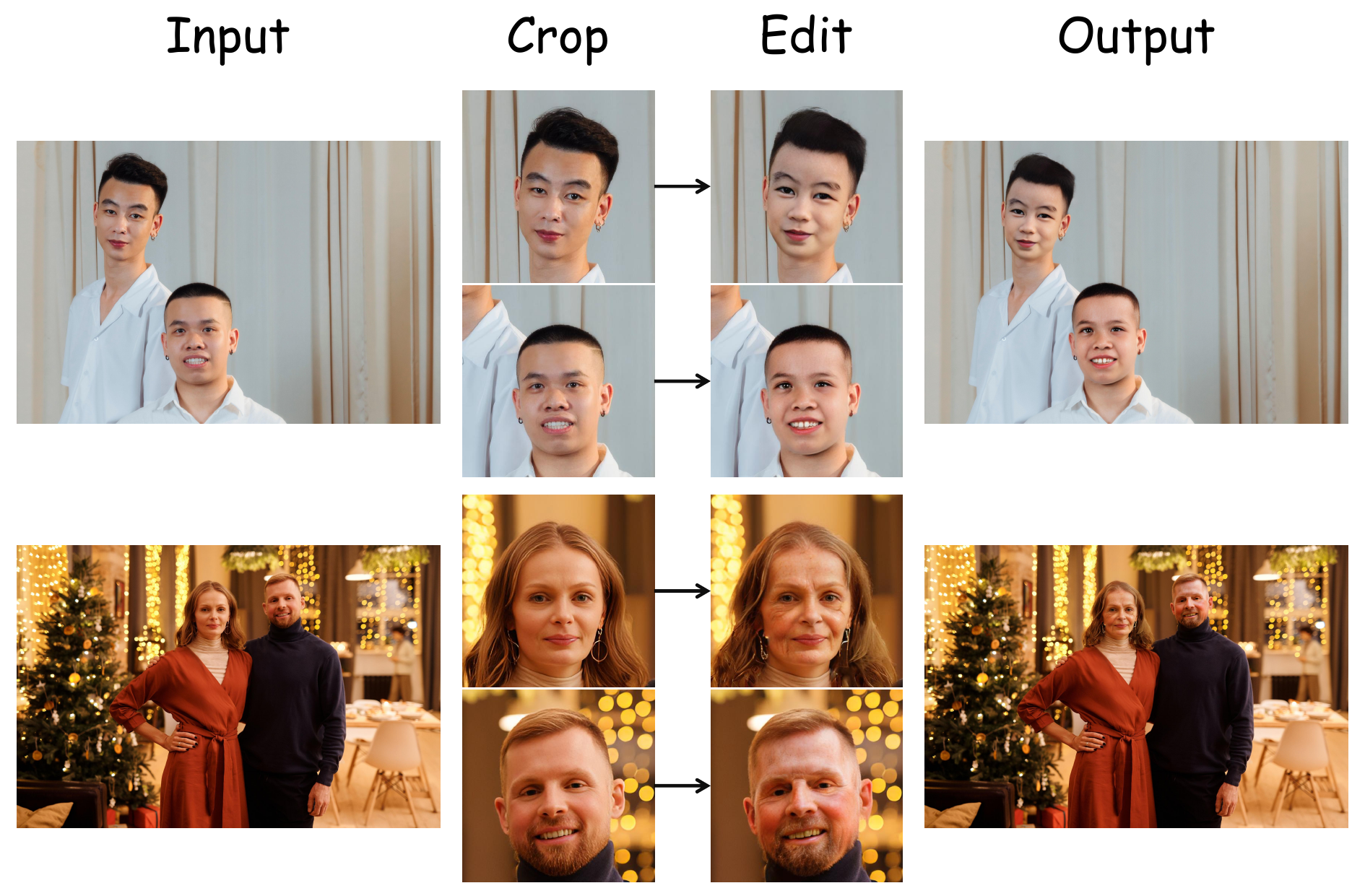}
\caption{\rev{Practical multi-subject facial age editing through a crop-edit-composite pipeline. Each face is localized, cropped, and edited independently before being composited back into the original full-scene image.}}
\label{fig:multiface_pipeline}
\vspace{-10pt}
\end{figure}

\subsubsection{\rev{Evaluation under Diverse Conditions and Failure Cases}}

\begin{revblock}
TimeMachine performs localized single-face editing, associating each model invocation with one identity and one target age. For multi-subject scenes, we evaluate a crop-edit-composite extension that localizes each face, applies the same trained model with an independently assigned target age, and composites the edited faces back into the original scene. This per-face procedure requires no architectural modification or additional training.

As shown in \figref{fig:multiface_pipeline}, the first example applies rejuvenation to both subjects, whereas the second applies aging to both subjects.
The edited faces retain recognizable subject-specific characteristics and are composited back into their original scenes while preserving the surrounding non-facial regions and scene composition.

\textbf{Diverse input conditions.}
Beyond relatively frontal portraits with simple visual conditions, we evaluate representative profile views, exaggerated expressions, complex backgrounds, makeup, low-light or low-quality inputs, and artistic styles.
For each input, TimeMachine generates results at target ages of 10, 40, and 70, as shown in \figref{fig:diverse_input_conditions}.
\end{revblock}

\begin{figure*}[!t]
\centering

\begin{minipage}[t]{0.49\textwidth}
\centering
\includegraphics[width=\linewidth]{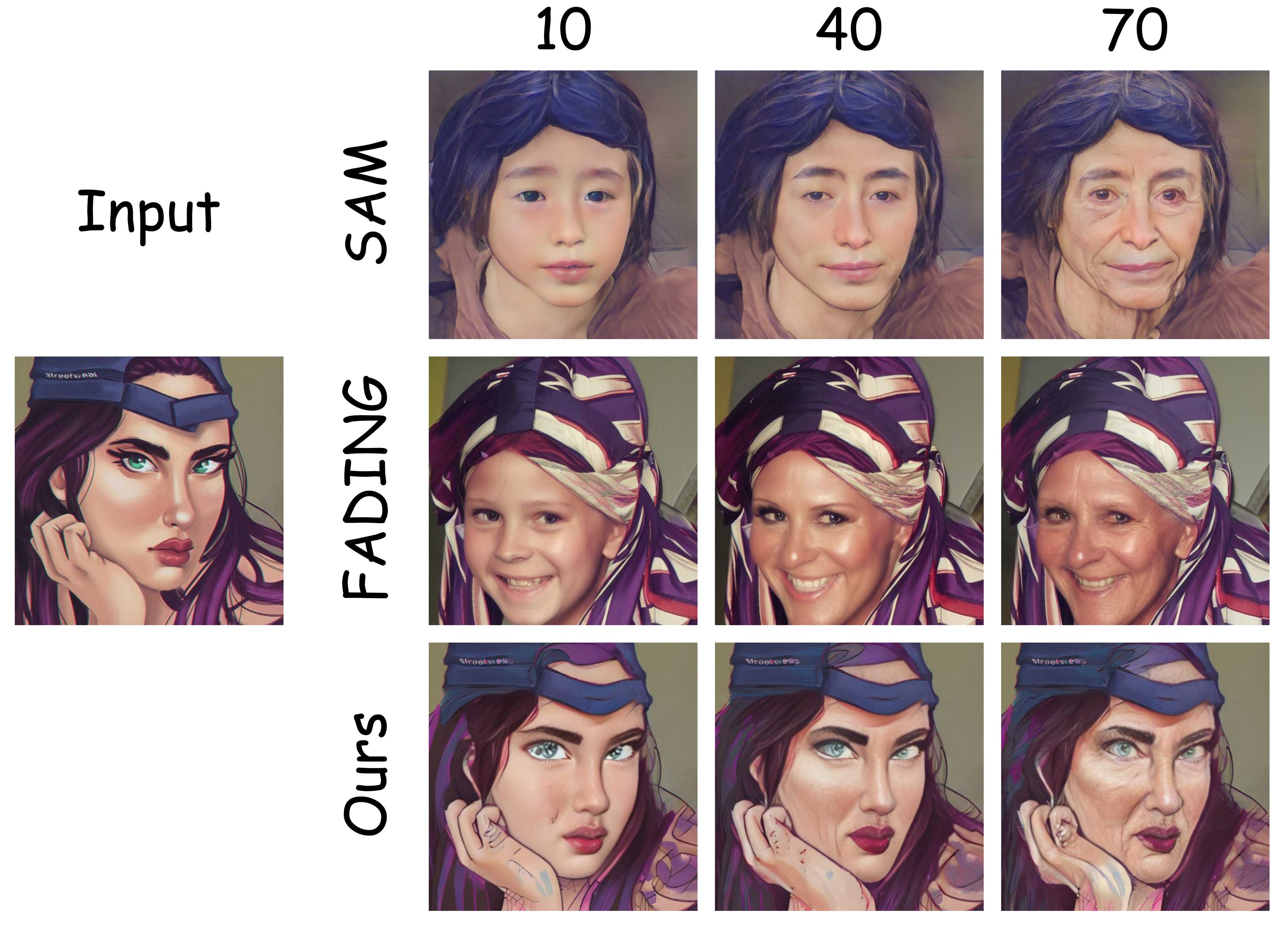}\\[-0.25em]
\textbf{(a) Stylized illustration}
\end{minipage}
\hfill
\begin{minipage}[t]{0.49\textwidth}
\centering
\includegraphics[width=\linewidth]{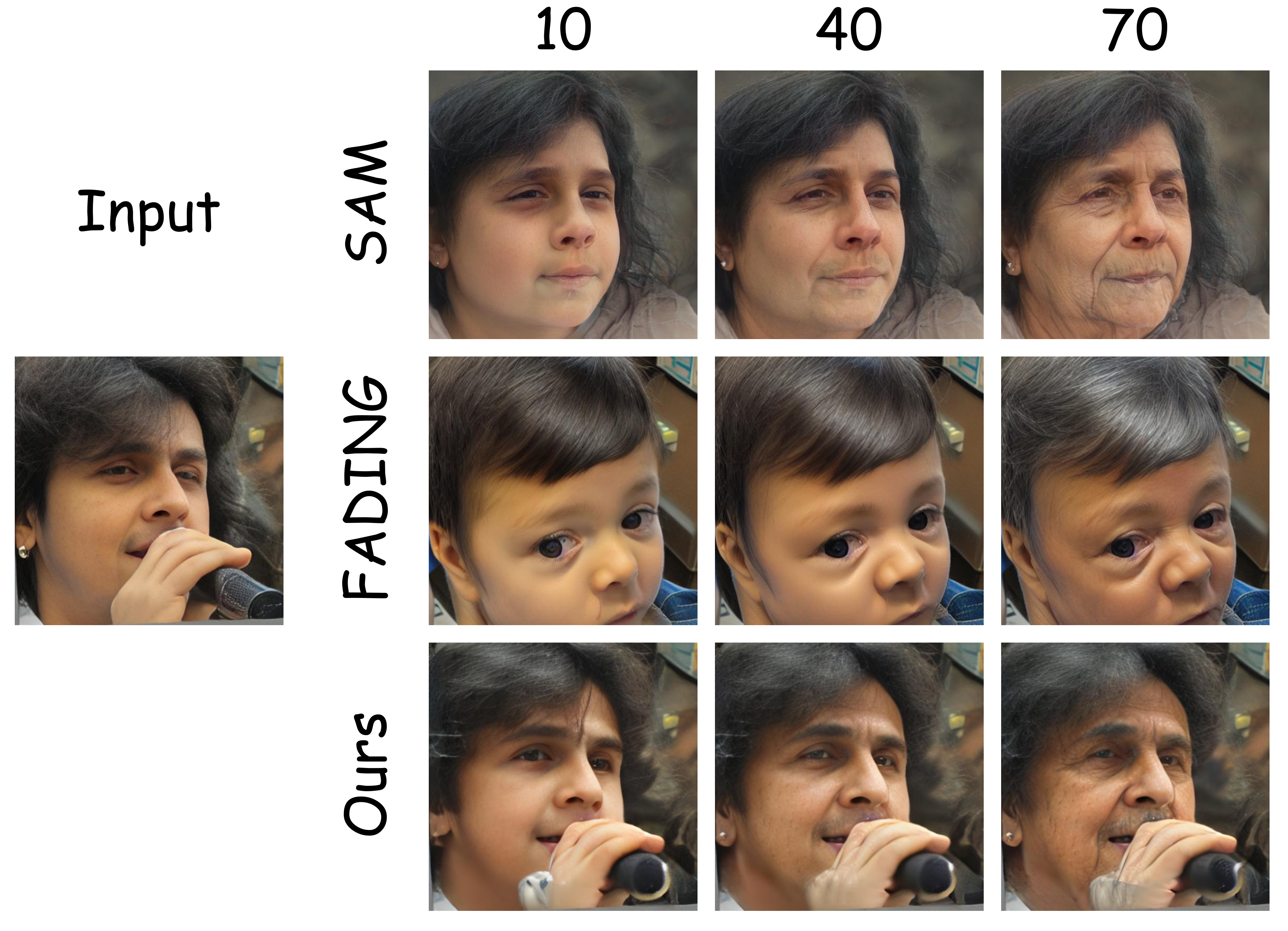}\\[-0.25em]
\textbf{(b) Microphone occlusion}
\end{minipage}

\caption{\rev{Representative challenging cases illustrating residual failure modes under a stylized input and a foreground occlusion. Each input is edited toward target ages of 10, 40, and 70 using SAM, FADING, and TimeMachine.}}
\label{fig:failure_cases}
\end{figure*}

\begin{revblock}
Across the evaluated profile and exaggerated-expression examples, TimeMachine introduces progressive age-related changes while retaining the principal viewing direction, expression, and facial structure.
For inputs with complex backgrounds, makeup, and challenging illumination or input quality, the corresponding non-age visual characteristics remain largely consistent throughout the editing sequence.
The representative artistic inputs also retain their principal visual characteristics while exhibiting recognizable age progression.

\textbf{Failure cases and limitations.}
We further evaluate one strongly stylized portrait and one foreground-occlusion case at target ages of 10, 40, and 70 using SAM, FADING, and TimeMachine.
Across both cases in \figref{fig:failure_cases}, TimeMachine retains more of the input-specific facial structure and relevant artistic or foreground content, while residual errors remain in style fidelity, occlusion boundaries, and extreme rejuvenation.
\end{revblock}

\begin{revblock}
For the stylized illustration in \figref{fig:failure_cases}(a), TimeMachine retains the hand-on-face pose, facial orientation, headwear, and principal color composition while introducing progressive age-related changes, whereas SAM and FADING exhibit larger changes in the source-specific facial appearance or composition.
Nevertheless, the exact linework and local geometry of the headwear and hand are not fully preserved; the youngest result remains visually older than the requested age, and the oldest result develops more realistic skin texture.

For the microphone occlusion in \figref{fig:failure_cases}(b), TimeMachine retains the input identity, pose, and microphone more consistently while producing visible age progression; even so, extreme rejuvenation remains imperfect and small geometric changes appear around the hand, mouth, and microphone boundary.

Overall, TimeMachine preserves more of the principal subject and input-specific structure than the compared methods, while performance degrades for strongly out-of-distribution styles, complex occlusion boundaries, and extreme target ages.
These limitations arise because the current framework lacks explicit foreground-layer decomposition, occlusion masks, and style-preservation objectives.
\end{revblock}

\subsection{Ablation Studies}
\begin{revblock}
To isolate the effects of the dedicated age and identity conditions from ACG, all variants in this branch ablation are trained without ACG and evaluated using the same 500-image protocol.
``W/O Age Conditions'' removes both the averaged age-codebook branch and the numerical age-text branch, whereas ``W/O ID Branch'' removes only the Antelopev2-based identity branch.
\end{revblock}

\begin{figure*}[htb]
    \centering
    \includegraphics[width=\textwidth]{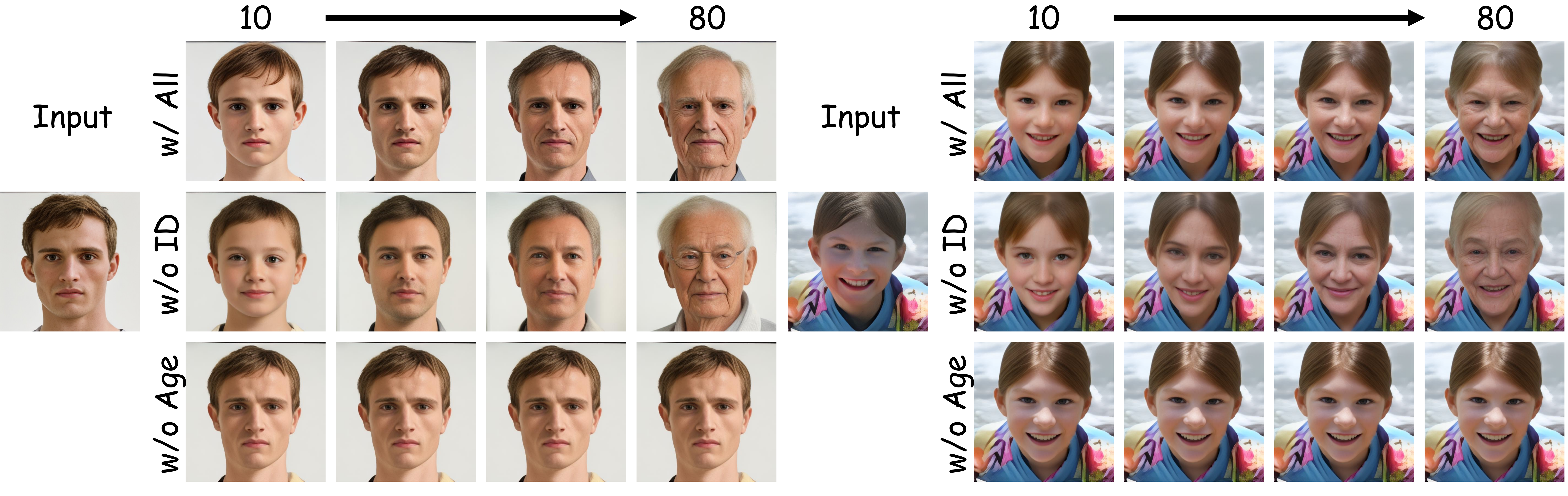}
    \caption{Ablation study on different condition controls. We demonstrate the performance of TimeMachine with different condition control at ages of 10, 30, 50, and 80. }
    \label{fig:ablation_crop}
    \vspace{-15pt}
\end{figure*}

The qualitative effects of these removals are illustrated in \figref{fig:ablation_crop}, and the quantitative results are reported in \tabref{tab:ablation1}. \rev{In the displayed examples, removing the dedicated age conditions substantially weakens the progression toward the requested target ages, whereas removing the ID branch introduces pronounced changes to the input-specific facial structure. The qualitative differences therefore mirror the age-accuracy and identity-similarity changes observed in the quantitative ablation.}

\begin{table}[h]
\centering

\small
\caption{\rev{Ablation of the dedicated age and identity conditions on the 500-image CelebA-HQ test set. All variants are trained without ACG.}}
\label{tab:ablation1}
\setlength{\tabcolsep}{3.5pt}
\resizebox{\linewidth}{!}{%
\begin{tabular}{@{}lccc@{}}
\toprule
Variant & MiVOLO MAE $\downarrow$ & MIVIA MAE $\downarrow$ & AdaFace Sim. $\uparrow$ \\ \midrule
Multi-CA (W/O ACG) & \textbf{3.242} & \textbf{5.742} & \textbf{0.679} \\
W/O Age Conditions & 20.609 & 19.569 & 0.676 \\
W/O ID Branch & 4.298 & 5.875 & 0.092 \\ \bottomrule
\end{tabular}
}
\end{table}

Removing the two dedicated age conditions increases \rev{MiVOLO and MIVIA MAE from 3.242 and 5.742 to 20.609 and 19.569, respectively}, while AdaFace similarity remains close \rev{(0.676 versus 0.679)}.
Removing the ID branch retains substantially better age accuracy \rev{at 4.298 and 5.875}, but reduces AdaFace similarity \rev{from 0.679 to 0.092}.

\rev{These results support the complementary primary functions of the dedicated age and ID conditions and provide evidence of branch-wise functional decoupling.}

\subsection{\rev{Multi-CA Design Analysis}}

\begin{revblock}
Under the same protocol without ACG, we compare Multi-CA with KV Concatenation, which uses one shared attention normalization, and Dynamic Gated Fusion, which predicts weights for independently normalized branch outputs; all unrelated settings remain unchanged.
\end{revblock}

\begin{table}[h]

\centering
\small
\caption{\rev{Attention-fusion comparison without ACG.}}
\label{tab:attention_fusion}
\setlength{\tabcolsep}{3.5pt}
\resizebox{\linewidth}{!}{%
\begin{tabular}{@{}lccc@{}}
\toprule
Method & MiVOLO MAE $\downarrow$ & MIVIA MAE $\downarrow$ & AdaFace Sim. $\uparrow$ \\
\midrule
KV Concatenation & 3.532 & 6.456 & 0.674 \\
Dynamic Gated Fusion & 3.362 & 6.227 & 0.676 \\
Multi-CA (Ours) & \textbf{3.242} & \textbf{5.742} & \textbf{0.679} \\
\bottomrule
\end{tabular}%
}
\end{table}

\begin{revblock}
As shown in \tabref{tab:attention_fusion}, Multi-CA achieves the best age accuracy and identity similarity, outperforming both shared-softmax KV Concatenation and Dynamic Gated Fusion under the matched protocol. \rev{The larger degradation of KV Concatenation is consistent with interference among heterogeneous conditions when they compete within a shared attention normalization. Dynamic Gated Fusion improves over KV Concatenation but remains inferior to Multi-CA on all three metrics, indicating that its additional content- and timestep-dependent weighting does not provide a measurable benefit under this protocol. These results support independent branch normalization followed by fixed fusion as a simple and effective design for coordinating the age and identity conditions.}
\end{revblock}

\begin{table}[h]

\centering
\small
\caption{\rev{Inference-time adapter composition versus jointly optimized Multi-CA.}}
\label{tab:multi_adapter_composition}
\setlength{\tabcolsep}{3.5pt}
\resizebox{\linewidth}{!}{%
\begin{tabular}{@{}lccc@{}}
\toprule
Method & MiVOLO MAE $\downarrow$ & MIVIA MAE $\downarrow$ & AdaFace Sim. $\uparrow$ \\
\midrule
Adapter composition & 6.409 & 9.813 & 0.392 \\
Multi-CA (W/O ACG) & 3.242 & 5.742 & 0.679 \\
Full TimeMachine & \textbf{3.081} & \textbf{5.565} & \textbf{0.682} \\
\bottomrule
\end{tabular}%
}
\end{table}

\begin{revblock}
We form an inference-time composition baseline by combining the official IP-Adapter-FaceID checkpoint~\cite{ye2023ip} with the independently trained age-condition module while keeping both modules fixed. The composition is evaluated on the same 500 images and seven target ages.
\end{revblock}

\begin{revblock}
As shown in \tabref{tab:multi_adapter_composition}, jointly optimized Multi-CA substantially improves both age accuracy and identity similarity over the inference-time composition baseline. \rev{Importantly, this performance gap is already substantial when the composition baseline is compared with Multi-CA without ACG, showing that it cannot be attributed solely to the auxiliary ACG objective. Within this controlled implementation, the results support jointly optimizing the specialized identity and age branches for the same editing objective instead of combining independently trained modules only at inference time. The additional improvement from Multi-CA without ACG to the full model separately reflects the contribution of ACG.}
\end{revblock}

\subsection{\rev{Effectiveness of ACG}}

We evaluate the effect of ACG \rev{using the 500-image CelebA-HQ protocol} and report results \rev{under both MiVOLO and the independent MIVIA estimator}.

\begin{table}[h]
\centering
\small
\caption{\rev{Effectiveness of the training-time ACG objective.}}
\label{tab:acg_ablation}
\setlength{\tabcolsep}{4pt}
\resizebox{\linewidth}{!}{%
\begin{tabular}{lccc}
\toprule
Variant
& MiVOLO MAE $\downarrow$
& MIVIA MAE $\downarrow$
& AdaFace Sim. $\uparrow$ \\
\midrule
Full model
& \textbf{3.081}
& \textbf{5.565}
& \textbf{0.682} \\
Without ACG
& 3.242
& 5.742
& 0.679 \\
\bottomrule
\end{tabular}%
}
\end{table}

As shown in \tabref{tab:acg_ablation}, ACG improves age accuracy \rev{under both evaluators} while maintaining identity similarity.
\rev{The corresponding MIVIA gain shows that the training objective also improves age accuracy measured by an evaluation-only estimator. A timestep-wise analysis on 150 age-stratified external CelebA-HQ images, combining a one-dimensional Isomap visualization with two structural metrics computed in the original 384-dimensional embedding space, is provided in Sec.~III-A of the supplementary material.}

\subsection{Analysis of Condition Control} \label{subsec:cond}

\begin{figure}[h]
    \centering
    \includegraphics[width=\linewidth]{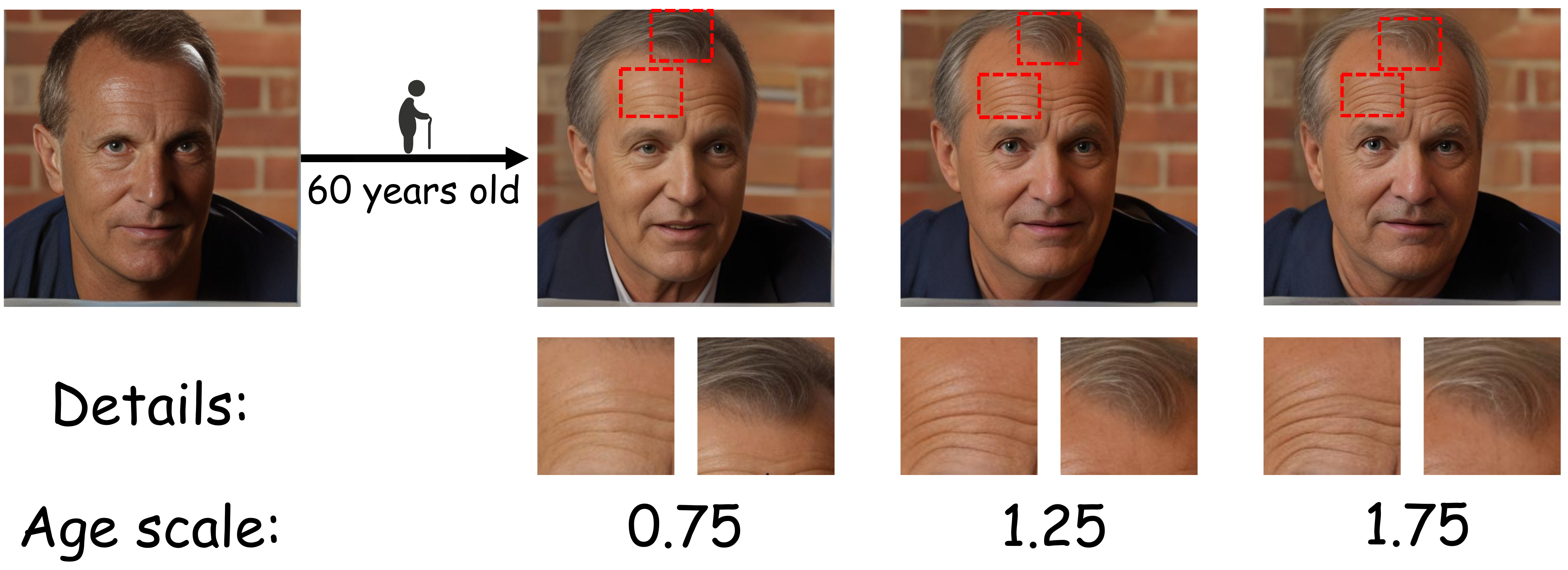}
    \caption{Visualization of age condition control by age scale.}
    \label{fig:condition_control}
\end{figure}

The diagnostic scale sweep in \figref{fig:condition_control} examines how the inference-time age-branch coefficient affects the strength of age-related changes.
Increasing this coefficient produces progressively more visible age cues in the displayed examples, including changes in facial texture, wrinkles, and hair appearance.
This sweep illustrates inference-time user control, \rev{while the same fixed scale configuration specified in the experimental setup is used for every sample in the quantitative comparisons.}

These observations show that the age-conditioned branch provides a controllable adjustment of age-transformation strength \rev{while the dedicated identity condition remains active throughout editing.}

\subsection{Analysis of Branch-Specific Attention}
\label{subsec:multi}

\begin{figure}[h]
    \centering
    \includegraphics[width=\linewidth]{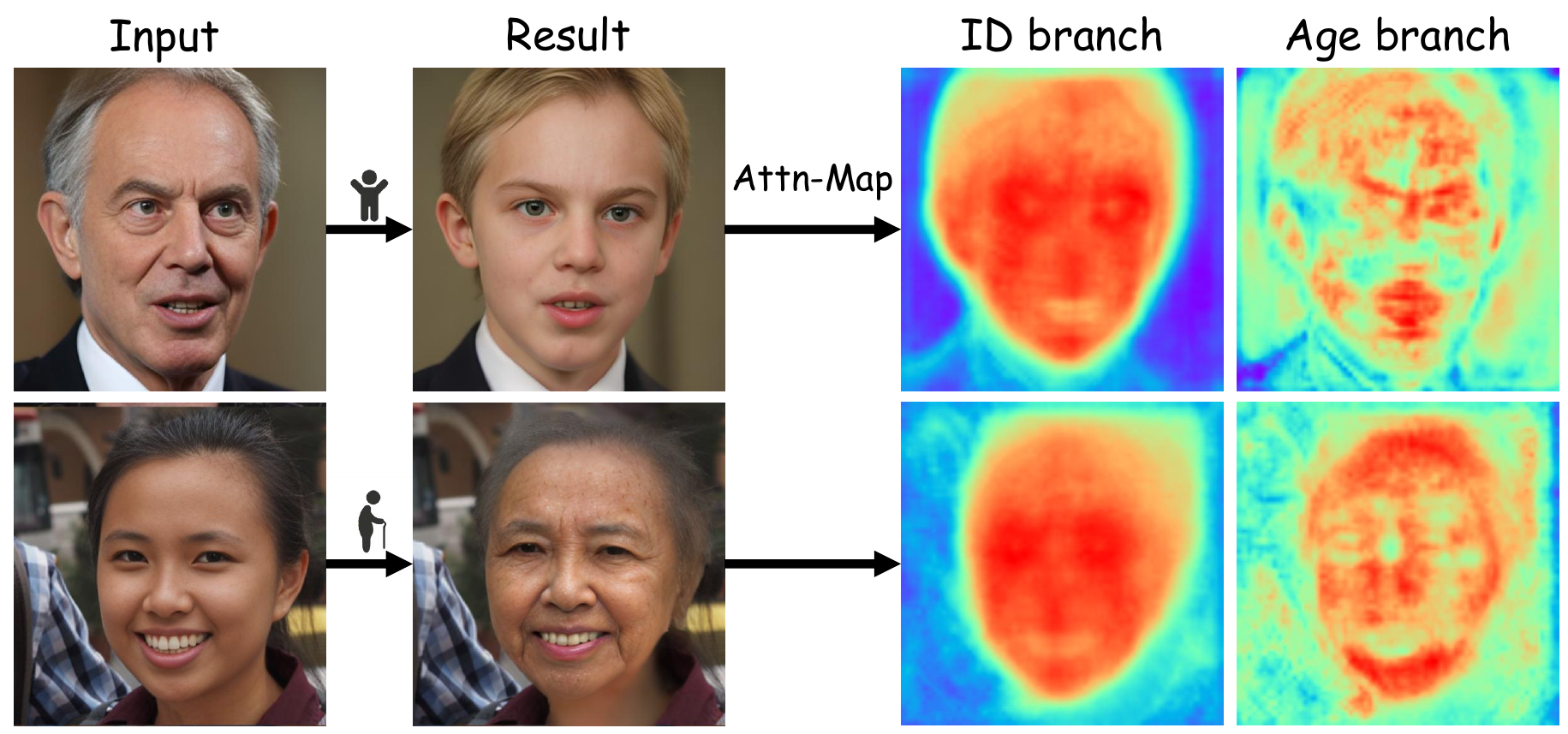}
    \caption{Visualization of branch-specific attention maps.}
    \label{fig:attention_map}
\end{figure}

In Multi-CA, the \rev{original text, numerical age-text, identity, and averaged age-codebook conditions} are processed through separate cross-attention branches \rev{with condition-specific representations and attention parameters}.
To examine the primary attention patterns learned by the dedicated identity and age branches, we visualize their cross-attention maps in \figref{fig:attention_map}.

\textbf{Identity Branch}: The visualized maps emphasize comparatively global facial structures, including face shape, consistent with the branch's identity-preserving role.

\textbf{Age Branch}: The visualized maps place stronger emphasis on local age-sensitive regions.
For younger targets, attention is directed toward regions around the hair, eyes, and mouth; for older targets, it shifts toward regions containing forehead, cheek, and jawline aging cues.

\rev{These maps characterize the primary spatial patterns of the identity and age branches, while the controlled ablations quantify their functional contributions. Together, the two analyses support the branch-specific roles of the dedicated identity and age conditions.}

\subsection{{User Study}} \label{subsec:user-study}

We conduct a \rev{method-blind} user study comparing TimeMachine with SAM \cite{alaluf2021only} and FADING \cite{chen2023face}.
\begin{revblock}
We recruited 50 participants and retained 49 complete responses; one incomplete submission was excluded, which was the only exclusion criterion.
Each valid participant evaluated 20 pairwise comparisons, resulting in 980 completed trials and 2,940 criterion-level judgments.

The study used a pool of 40 comparison tasks, comprising 20 TimeMachine-versus-SAM tasks and 20 TimeMachine-versus-FADING tasks.
Source images were randomly drawn from a combined pool containing CelebA-HQ \cite{karras2017progressive}, MORPH \cite{ricanek2006morph}, AgeDB \cite{AGEDB8014984}, FFHQ \cite{karras2019style}, and CACD \cite{chen2014cross}, and target ages were randomly selected from the range of 10--80 years.
Task sampling was independent of the output quality of all evaluated methods.
Each participant was randomly assigned 10 comparisons with SAM and 10 with FADING.

Within each trial, TimeMachine and the compared method used the same input image and target age. Method names were hidden, left/right output positions were randomized, and participants selected Output A, ``Similar,'' or Output B according to three criteria: target-age accuracy, identity preservation, and overall visual quality.
Participants were informed of the study purpose and participated voluntarily and anonymously; no personally identifying or sensitive information was collected.
The blind evaluation interface and trial layout are shown in Sec.~IV of the supplementary material.
\end{revblock}

\begin{table}[h]
\centering
\caption{\rev{Results of the blind user study. All values are percentages.}}
\label{tab:user_study_results}
\setlength{\tabcolsep}{3pt}
\renewcommand{\arraystretch}{1.08}
\resizebox{\linewidth}{!}{%
\begin{tabular}{llcccc}
\toprule
\multirow{2}{*}{Method} &
\multirow{2}{*}{Criterion} &
\multicolumn{3}{c}{Preference} &
\multirow{2}{*}{Score (95\% CI)} \\
\cmidrule(lr){3-5}
& & TM & Tie & Comp. & \\
\midrule
\multirow{3}{*}{SAM}
& Age       & 74.9 & 13.3 & 11.8 & 81.5 (78.6--84.5) \\
& Identity  & 70.2 & 16.1 & 13.7 & 78.3 (75.0--81.5) \\
& Quality   & 72.2 & 13.1 & 14.7 & 78.8 (75.4--82.0) \\
\midrule
\multirow{3}{*}{FADING}
& Age       & 73.9 & 13.9 & 12.2 & 80.8 (77.8--83.9) \\
& Identity  & 72.4 & 15.7 & 11.8 & 80.3 (77.0--83.5) \\
& Quality   & 70.2 & 15.5 & 14.3 & 78.0 (75.3--80.6) \\
\bottomrule
\end{tabular}%
}
\end{table}

\begin{revblock}
For statistical analysis, a preference for TimeMachine, ``Similar,'' and the compared method was assigned a score of 1, 0.5, and 0, respectively.
Preference scores were first calculated for each participant, and 95\% confidence intervals were estimated using 100,000 participant-level bootstrap resamples.
Statistical significance against the no-preference score of 0.5 was assessed using a two-sided participant-level Wilcoxon signed-rank test, with a Bonferroni-corrected threshold of $0.05/6=0.0083$ for the six criterion-by-method comparisons.
\end{revblock}

\begin{revblock}
In \tabref{tab:user_study_results}, TM and Comp. denote TimeMachine and the compared method, respectively.
TimeMachine is directly preferred in 70.2\%--74.9\% of the judgments across the three criteria, whereas the compared methods are preferred in only 11.8\%--14.7\%.
After accounting for ties with a score of 0.5, the participant-level preference scores range from 78.0\% to 81.5\%.
All 95\% confidence intervals remain entirely above the no-preference level of 50\%, and all six comparisons achieve $p<0.001$, remaining significant under the Bonferroni-corrected threshold.
These results show that the perceptual advantage of TimeMachine is consistently observed in target-age accuracy, identity preservation, and visual quality.
\end{revblock}

\section{Conclusion}

We present \textbf{TimeMachine}, \rev{a diffusion-based facial age editor combining the HFFA training collection, an averaged numerical-age codebook, dedicated identity and age conditions, branch-wise Multi-CA, and training-only ACG.}
\rev{Across five public face datasets,} it achieves superior age accuracy and identity preservation relative to the evaluated baselines, \rev{with consistent evidence from five age evaluators, FairFace-assigned Groups 1--4, and the blind user study.}
\rev{Together, the age codebook and dedicated branches balance numerical-age control and identity preservation, while ACG adds training-only supervision without changing inference.}
\rev{The model edits one localized face per invocation; exact preservation of strongly out-of-distribution styles and occlusion boundaries remains challenging, while multi-subject scenes are supported through the evaluated crop-edit-composite extension.}


\begin{IEEEbiography}[{\includegraphics[width=1in,height=1.25in,clip,keepaspectratio]{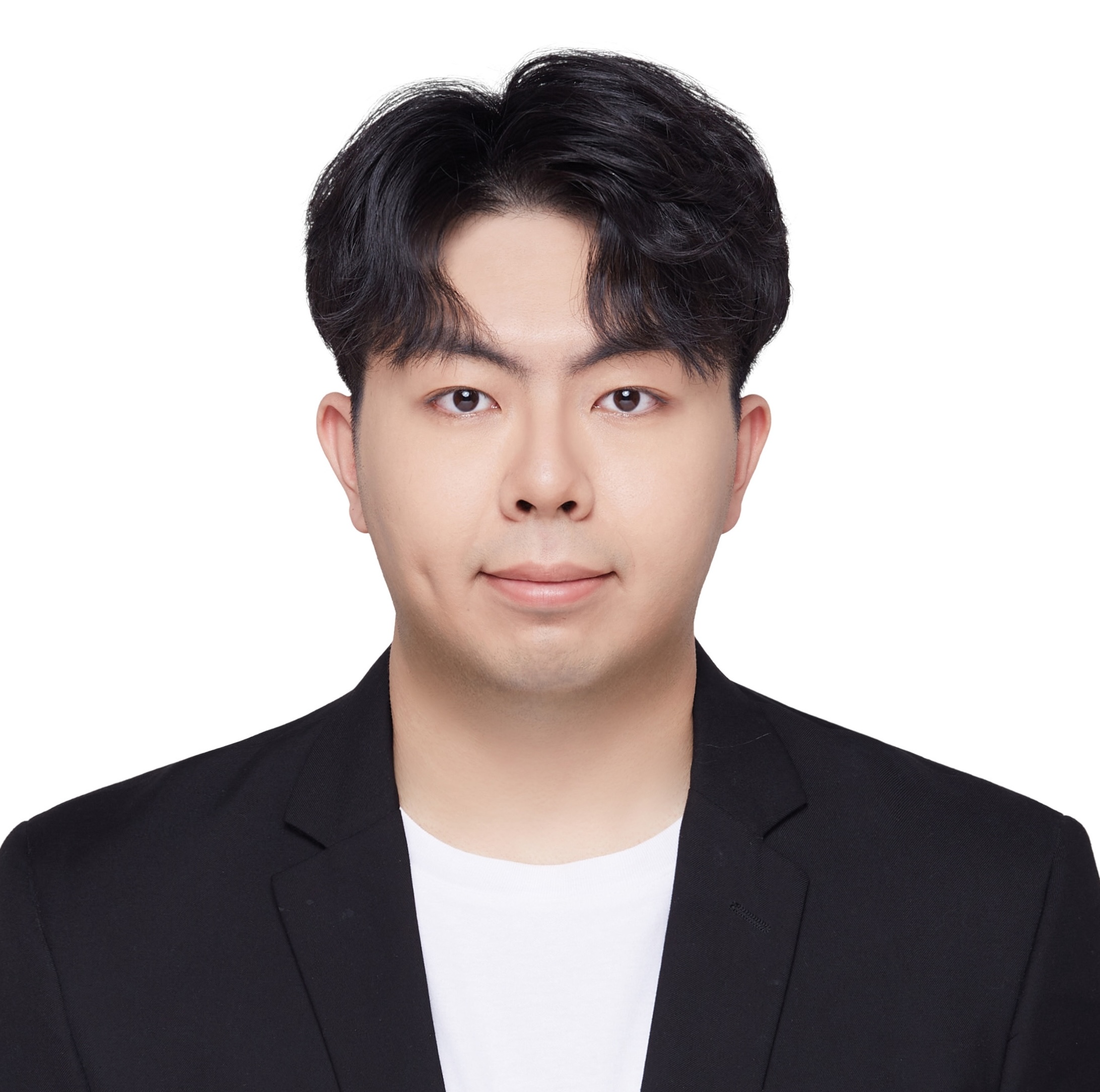}}]{Yilin Mi}
received the B.S. degree in Computer Science and Technology from Nankai University, Tianjin, China, in 2025, where he is currently pursuing the M.S. degree with the Media Computing Laboratory, Nankai University, supervised by Prof. Xiuli Shao and Prof. Chongyi Li. 
His research interests include computer vision, with a particular focus on interactive image editing.
\end{IEEEbiography}

\vspace{-1cm}
\begin{IEEEbiography}[{\includegraphics[width=1in,height=1.25in,clip,keepaspectratio]{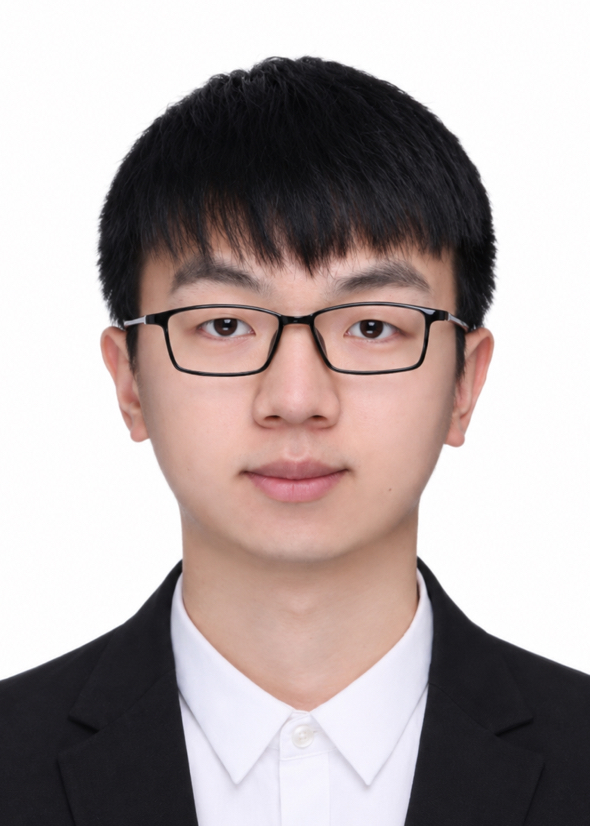}}]{Qixin Yan}
received the master’s degree from Tianjin University, Tianjin, China, in 2022. He is currently a Researcher with Weixin, Tencent. His research interests include generative models and AI-generated content (AIGC).
\end{IEEEbiography}

\vspace{-1cm}
\begin{IEEEbiography}[{\includegraphics[width=1in,height=1.25in,clip,keepaspectratio]{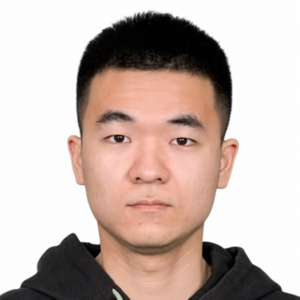}}]{Zheng-Peng Duan}
received the B.S. degree from Xidian University, China, in 2022. He is currently pursuing the Ph.D. degree with the Media Computing Laboratory, Nankai University, supervised by Prof. Chongyi Li and Prof. Ming-Ming Cheng. His research interests include image restoration and interactive editing.
\end{IEEEbiography}

\vspace{-1cm}
\begin{IEEEbiography}
[{\includegraphics[width=1in,height=1.25in,clip,keepaspectratio]{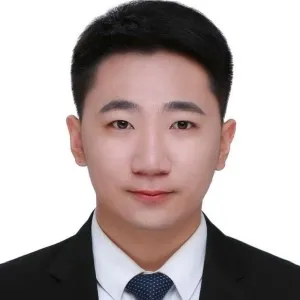}}]{Chunle Guo}
(Member, IEEE) received the Ph.D. degree from Tianjin University, China, under the supervision of Prof. Ji-Chang Guo. He was a Visiting Ph.D. Student with the School of Electronic Engineering and Computer Science, Queen Mary University of London (QMUL), U.K. He was a Research Associate with the Department of Computer Science, City University of Hong Kong (CityUHK). He was a Postdoctoral Researcher with Nankai University, under the guidance of Prof. Ming-Ming Cheng. He is currently an Associate Professor with Nankai University. 
His current research interests include image processing, computer vision, and deep learning, particularly in the domains of image restoration and enhancement.
\end{IEEEbiography}

\vspace{-1cm}
\begin{IEEEbiography}[{\includegraphics[width=1in,height=1.25in,clip,keepaspectratio]{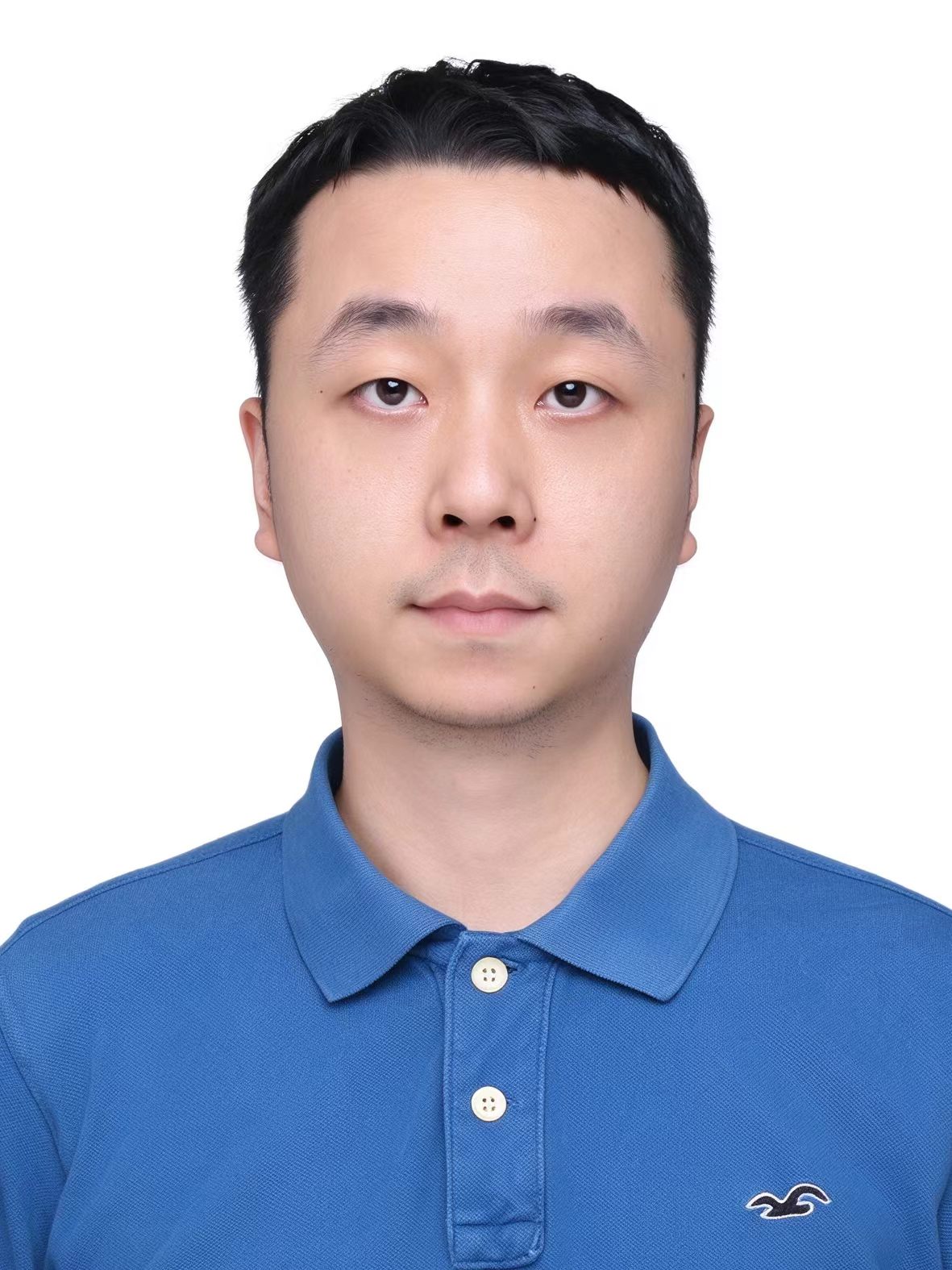}}]{Hubery Yin}
received the master’s degree in computer science from Zhejiang University in 2017. He is currently a Researcher with Weixin, Tencent. Research interests include generative models.
\end{IEEEbiography}

\vspace{-1cm}
\begin{IEEEbiography}[{\includegraphics[width=1in,height=1.25in,clip,keepaspectratio]{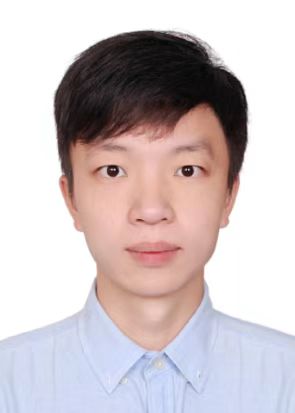}}]{Hao Liu}
received his B.S. and M.S. degrees in Computer Science and Technology from Sichuan University, Chengdu, China, in 2012 and 2015, respectively. He is currently a Senior Researcher and Team Lead at WeChat Vision Center, Tencent. His research interests include computer vision, graphics, generative models, and AIGC.
\end{IEEEbiography}

\vspace{-1cm}
\begin{IEEEbiography}[{\includegraphics[width=1in,height=1.25in,clip,keepaspectratio]{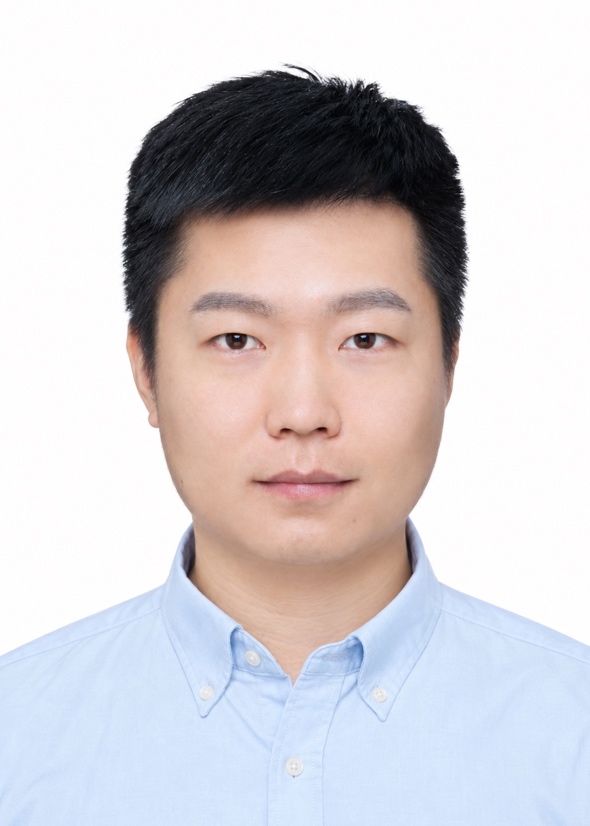}}]{Chen Li}
is a senior researcher at Tencent and leads the perception and generation at WeChat vision team. He obtained his Ph.D. from Zhejiang University in 2017 and worked at DJI Innovation and Tencent YouTu Lab. In 2020, he joined the Tencent WeChat.
\end{IEEEbiography}

\vspace{-1cm}
\begin{IEEEbiography}[{\includegraphics[width=1in,height=1.25in,clip,keepaspectratio]{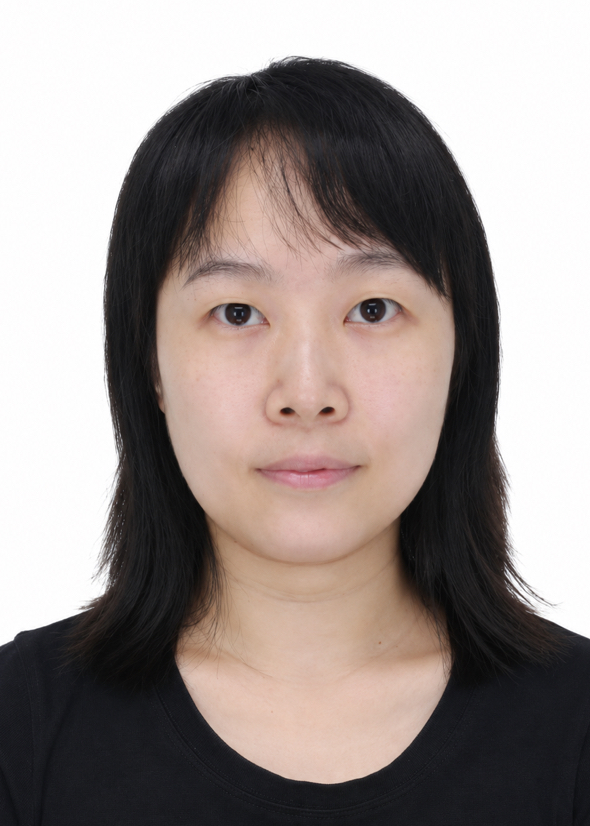}}]{Jing Lyu}
is currently leading the Visual and Multimedia Technology Teams with Weixin, Tencent. Her research interests focus on multimedia AI, including large models for visual generation and multimodal understanding, as well as terminal inference engines. She has directed the research and development of several foundational AI components, such as multimodal embedding, native video understanding, and image generation models, bridging advanced AI research with large-scale engineering deployment.
\end{IEEEbiography}

\vspace{-1cm}
\begin{IEEEbiography}
[{\includegraphics[width=1in,height=1.25in,clip,keepaspectratio]{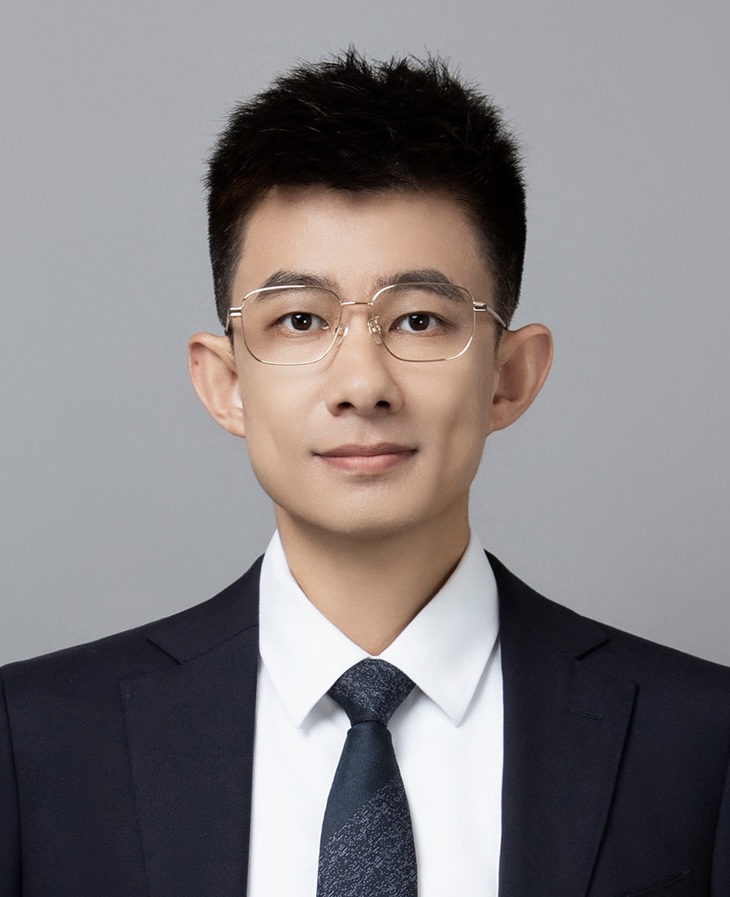}}]{Chongyi Li}
(Senior Member, IEEE) received the Ph.D. degree from the School of Electrical and Information Engineering, Tianjin University, Tianjin, China, in June 2018. From 2016 to 2017, he was a Joint Training Ph.D. Student with The Australian National University, Australia. He was a Research Fellow with the City University of Hong Kong and Nanyang Technological University from 2018 to 2021. He was a Research Assistant Professor with the School of Computer Science and Engineering, Nanyang Technological University, from 2021 to 2023. He is currently a Full Professor with the School of Computer Science, Nankai University, China. 
His current research interests
include computational imaging, generative vision, and visual intelligence.
\end{IEEEbiography}

\vspace{1cm}
\thanks{This paper has supplementary downloadable material available at
https://ieeexplore.ieee.org, provided by the authors. The material consists
of a PDF document containing additional experimental results and analyses.
Please contact \mbox{yilinmi@mail.nankai.edu.cn} for further questions
about this work.}

\vfill
\end{document}